%% file: 0_main.tex
\documentclass[journal]{IEEEtran}
\usepackage[compatibility=false]{caption}
\usepackage[switch]{lineno}
\usepackage{cite}
\usepackage{booktabs}
\usepackage{enumitem}
\usepackage{amsmath,amssymb,amsfonts}
\usepackage[noend]{algpseudocode}
\usepackage[ruled,linesnumbered]{algorithm2e}
\usepackage{algpseudocode}
\usepackage{multirow}
\usepackage{multicol}
\usepackage{amsmath}
\usepackage{array}
\usepackage{graphicx}
\usepackage{caption}
\usepackage{subcaption}
\usepackage{overpic}
\usepackage{textcomp}
\usepackage{xcolor}

\usepackage{stfloats}
\usepackage{textcomp}
\usepackage{pifont}
\usepackage[colorlinks,
            linkcolor=blue,
            anchorcolor=black,
            citecolor=blue]{hyperref}

\begin{document}
%

\title{A Multi-Agent Reinforcement Learning Approach for Cooperative Air–Ground–Human Crowdsensing in Emergency Rescue}


\author{
        Wenhao~Lu,\IEEEmembership{}
        Zhengqiu~Zhu,\IEEEmembership{}
        Yong~Zhao,\IEEEmembership{}
        Yonglin~Tian,\IEEEmembership{~Member,~IEEE,}
        Junjie~Zeng,\IEEEmembership{}
        Jun~Zhang,\IEEEmembership{}
        Zhong~Liu,\IEEEmembership{~Member,~IEEE,}
        and Fei-Yue~Wang,\IEEEmembership{~Fellow,~IEEE}

\thanks{Wenhao Lu, Zhengqiu Zhu, Yong Zhao, Junjie Zeng, Jun Zhang, and Zhong Liu are with the College of Systems Engineering, National University of Defense Technology, Changsha 410073, Hunan Province, China. 
(e-mail:~luwenhao20@nudt.edu.cn;~zhuzhengqiu12@nudt.edu.cn;~zhaoyong15@nudt.edu.cn;~zengjunjie13@nudt.edu.cn;~zhangjun@nudt.edu.cn;~phillipliu@263.net).}
\thanks{Yongling Tian and Fei-Yue Wang are with the State Key Laboratory of Multimodal Artificial Intelligence Systems, Institute of Automation, Chinese Academy of Sciences, Beijing 100190, China 
(e-mail:~yonglin.tian@ia.ac.cn;~feiyue@ieee.org).}
}


\maketitle

\begin{abstract}
Mobile crowdsensing is evolving beyond traditional human-centric models by integrating heterogeneous entities like unmanned aerial vehicles (UAVs) and unmanned ground vehicles (UGVs). Optimizing task allocation among these diverse agents is critical, particularly in challenging emergency rescue scenarios characterized by complex environments, limited communication, and partial observability. 
This paper tackles the Heterogeneous-Entity Collaborative-Sensing Task Allocation (HECTA) problem specifically for emergency rescue, considering humans, UAVs, and UGVs. 
We introduce a novel ``Hard-Cooperative'' policy where UGVs prioritize recharging low-battery UAVs, alongside performing their sensing tasks. The primary objective is maximizing the task completion rate (TCR) under strict time constraints. We rigorously formulate this NP-hard problem as a decentralized partially observable Markov decision process (Dec-POMDP) to effectively handle sequential decision-making under uncertainty. To solve this, we propose HECTA4ER, a novel multi-agent reinforcement learning algorithm built upon a Centralized Training with Decentralized Execution architecture. HECTA4ER incorporates tailored designs, including specialized modules for complex feature extraction, utilization of action-observation history via hidden states, and a mixing network integrating global and local information, specifically addressing the challenges of partial observability. Furthermore, theoretical analysis confirms the algorithm's convergence properties. 
Extensive simulations demonstrate that HECTA4ER significantly outperforms baseline algorithms, achieving an average 18.42\% increase in TCR. Crucially, a real-world case study validates the algorithm's effectiveness and robustness in dynamic sensing scenarios, highlighting its strong potential for practical application in emergency response.
\end{abstract}


\input{1_introduction}
\input{2_related_work}

\input{3_model}

\input{4_method}
\input{5_experiment}
\input{6_discussion}

\input{7_conclusion}

\bibliographystyle{./IEEEtran}
\bibliography{IEEE_reference}


\section*{About the author}

\textbf{Wenhao Lu} received the B.S. degree in big data engineering from National University of Defense Technology, Changsha, China, in 2024. He is currently working toward the Ph.D. degree in control science and engineering. His current research focuses on the problems of mobile crowdsensing, deep reinforcement learning and embodied intelligence.

\textbf{Zhengqiu Zhu} received the Ph.D. degree in management science and engineering from National University of Defense Technology, Changsha, China, in 2023. He is currently a lecturer with the College of Systems Engineering, National University of Defense Technology, Changsha, China. His research interests include mobile crowdsensing, social computing, and LLM-based AI agents.

\medskip

\textbf{Yong Zhao} received the M.E. degree in control science and engineering from the National University of Defense Technology, Changsha, China, in 2021, where he is currently working toward the Ph.D. degree in management science and engineering. His current research focuses on crowdsensing, human-AI interaction, and embodied intelligence.

\medskip

\textbf{Yonglin Tian (Member, IEEE)} received the Ph.D. degree in control science and engineering from the University of Science and Technology of China, Hefei, China, in 2022. He is currently an Assistant Researcher with the Institute of Automation, Chinese Academy of Sciences, Beijing, China. His research interests include scenarios engineering, multimodal perception and parallel driving. He serves as an Associate Editor of IEEE Transactions on Intelligent Vehicles, Youth Editor of Research. He also served as the Guset Editor of IEEE Transactions on Systems, Man, and Cybernetics: Systems.

\medskip

\textbf{Junjie Zeng} received the M.E. degree in control
science and engineering from National University of Defense
Technology, Changsha, China, in 2019. He is currently a assistant researcher with the College of Systems Engineering, National University of Defense Technology, Changsha, China. His research interests include deep reinforcement learning, imitation learning and cognitive behavior modeling.

\medskip

\textbf{Jun Zhang} received the Ph.D. degree in control science and engineering from the National University of Defense Technology, Changsha, China, in 2008. She is currently a Professor with the College of Systems Engineering, National University of Defense Technology. She has published four books and more than 80 articles in top-tier journals and conferences. Her research interests include digital intelligence system engineering, artificial intelligence, and deep learning. 

\medskip

\textbf{Zhong Liu (Member, IEEE)} received the Ph.D. degree in system engineering and mathematics from the National University of Defense Technology, Changsha, China, in 2000. He is currently a professor of the National University of Defense Technology, Changsha, Hunan, China. He is also the head of Science and Technology Innovation Team, Ministry of Education. His main research interests include artificial general intelligence, deep reinforcement learning, and multi-agent systems. Prof. Liu is a member of the National Artificial Intelligence Strategy Advisory
Committee.

\medskip

\textbf{Fei-Yue Wang (Fellow, IEEE) } received the Ph.D. degree in computer and systems engineering from the Rensselaer Polytechnic Institute, USA, in 1990. He is the State Specially Appointed Expert and the Director of the State Key Laboratory for Management and Control of Complex Systems. He is also a Fellow of INCOSE, IFAC, ASME, and AAAS. His current research focuses on methods and applications for parallel systems, social computing, parallel intelligence, and knowledge automation. Detailed Bio. of Fei-Yue Wang can be found at www.compsys.ia.ac.cn/people/wangfeiyue.html.


\clearpage
\input{Appendix}

\end{document}

%% file: 1_introduction.tex
\section{Introduction}

\IEEEPARstart{W}{ith} the rapid adoption of mobile intelligent devices and the rise of crowd intelligence computing paradigms, mobile crowdsensing (MCS) \cite{Suhag2023,kim2022,liao2025drift} has witnessed extensive development. Unlike static sensing networks that rely on fixed sensors, MCS recruits mobile users to accomplish sensing tasks through crowdsourcing \cite{Hettiachchi2023,Ang2022}, offering advantages such as broad spatiotemporal coverage, low sensing costs, and flexible deployment. As a result,  MCS has been widely applied in areas such as smart transportation and environmental monitoring \cite{Suhag2023}. Apart from crowd workers, unmanned aerial vehicles (UAVs) and unmanned ground vehicles (UGVs) equipped with high-precision sensors also demonstrate significant potential for accomplishing sensing tasks. Therefore, some studies \cite{Zhou2024,Yang2023,Bai2024} have employed these unmanned vehicles to assist human workers in sensing activities. However, due to the energy constraints of UAVs and the limited sensing range of UGVs, relying on a single type of unmanned terminal makes it challenging to effectively complete sensing tasks. To address this issue, a collaborative sensing approach has been explored where UGVs carry UAVs and provide them with power replenishment \cite{Wang2023a,Zhao2024}. 

In the field of heterogeneous collaborative sensing, involving diverse entities such as crowd workers,  UGVs, and UAVs, task allocation represents a critical research problem. It centers on the optimal assignment of sensing tasks to available entities. 
Effective task allocation necessitates careful alignment between entity capabilities and specific task requirements. Furthermore, this process aims to optimize multiple objectives, typically including the minimization of critical operational parameters such as total sensing duration and associated costs, alongside the maximization of synergistic efficiency arising from the collaboration among heterogeneous sensing entities. 
This paper focuses on the task allocation problem for heterogeneous-entity collaborative sensing (HECTA) within emergency rescue scenarios \cite{Wang2007,yang2022parallel}. Compared to general applications, these scenarios introduce distinct challenges, including complex environments, severe infrastructure damage, and disrupted communication conditions. Although some studies have investigated the HECTA problem \cite{Wei2023,Yang2024,Guan2023a} in this scenario, the following issues remain to be addressed.

\begin{itemize}[leftmargin=*]
    \item Emergency rescue scenarios inherently involve complex environments and compromised communication conditions, leading to partially observable states \cite{Tang2024}. Operating with only partial observations presents a significant challenge for sensing entities, hindering their ability to accurately extract environmental features and formulate optimal decisions within such settings.
    
    \item The collaborative mode between humans, UAVs and UGVs needs to be carefully designed by comprehensively considering their distinct capabilities (e.g., sensing range, perception accuracy, and energy consumption), as well as time constraints, and task attributes. Additionally, the limited endurance of UAVs presents a challenge in designing UGV operation mechanisms to balance their own sensing tasks against supporting UAVs through battery recharging.
    
    \item Emergency rescue scenarios impose stringent requirements on the timeliness, accuracy and generalizability of task allocation algorithms \cite{Wang2023b,cui2025cooperative}. Existing methods often struggle to balance computational efficiency and allocation effectiveness, and typically exhibit limited generalization capabilities. Developing an algorithm that simultaneously achieves rapid solution speed, high task completion rates, and strong robustness remains a significant challenge.
\end{itemize}

\begin{figure}[!t]
\centering
\includegraphics[width=3.5in]{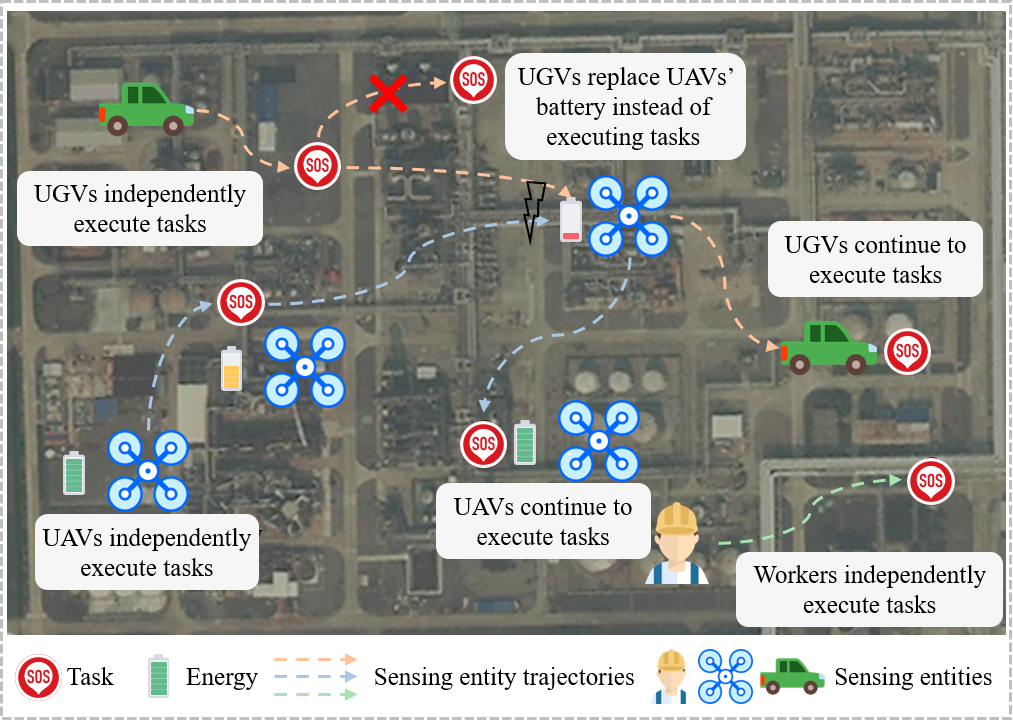}%
\caption{An illustrative process of HECTA in emergency rescue scenarios.}
\label{问题示意图}
\end{figure}

To address the aforementioned challenges, this paper investigates the HECTA problem involving human workers, UAVs and UGVs for emergency rescue. We define that humans, UAVs, and UGVs can perform distinct sensing tasks. Additionally, due to the limited endurance of UAVs, UGVs also operate under a ``Hard-Cooperative'' battery policy, prioritizing servicing low-battery UAVs over their own sensing. Further, we formulate this allocation problem as an optimization problem aimed at maximizing the task completion rate and prove its NP-hardness. The difficulty conventional methods face in balancing efficiency and effectiveness for NP-hard problems motivates our use of Multi-Agent Reinforcement Learning (MARL) \cite{10977657}. Furthermore, we prove that the agents operate in a fully cooperative manner, establishing a foundation for the subsequent MARL solution. Figure \ref{问题示意图} depicts an example HECTA process under our problem settings.

We formulate the problem as a decentralized partially observable Markov decision process (Dec-POMDP) to handle sequential allocation. To address the inherent partial observability, we use belief states, which represent the probability distribution over the current state given agent observations. We prove the belief state satisfies the Markov property, ensuring the model's mathematical soundness. Based on this, we introduce HECTA4ER, a MARL algorithm utilizing a Centralized Training with Decentralized Execution (CTDE) architecture. Its components include a convolutional module for extracting complex environmental feature, a decision-making module using hidden states to store action-observation history, and a mixing network processing both global and local information. To improve training performance, HECTA4ER incorporates an action filtering mechanism for sparse rewards and a novel replay buffer designed for sequential data.

The main contributions of this work are as follows:

\begin{itemize}[leftmargin=*]
    \item We formally define and formulate the HECTA problem by considering the condition of partially observable environmental states. Its NP-hard property as well as the fully cooperative characteristics between heterogeneous entities are proven. By modeling the problem as a Dec-POMDP, a solution is developed.
    \item We propose the MARL-based algorithm HECTA4ER to solve the task allocation problem. 
    The algorithm, which is well designed in terms of complex environmental information extraction, utilization of action-observation history, and dual processing of global and local information, can effectively handle emergency rescue scenarios under partial observation condition.
    \item We conduct extensive experiments using both simulation data and a real-world case study to evaluate the performance of HECTA4ER. Results from various simulated scenarios demonstrate that HECTA4ER achieves an average 18.42\% increase in task completion rate (TCR) compared to baseline methods. Notably, the real-world case study highlights the algorithm's strong robustness, showing its ability to maintain a higher TCR in dynamic sensing scenarios and indicating its potential for practical applications.
\end{itemize}

The remainder of this paper is organized as follows: Section \ref{2} reviews related work; Section \ref{3} presents system modeling; Section \ref{4} details the HECTA4ER algorithm; Section \ref{5} conducts extensive experiments and analyzes the results; Section \ref{6} discusses the findings, drawbacks and potential avenues; and Section \ref{7} concludes this work.


%% file: 2_related_work.tex
\section{Related work}\label{2}
\subsection{Changes in environment sensing approaches for emergency rescue}
In the early stages, environment sensing in emergency rescue scenarios primarily relied on static sensor networks and manual search efforts. Zeng \textit{et al.} \cite{Zeng2023} provided a comprehensive review of the role of sensors in urban Internet of Things (IoT) systems for disaster management. Wang \textit{et al.} \cite{Wang2024a} investigated optimal search and rescue route planning in earthquake scenarios, proposing a multi-point path selection algorithm for rescue operations. These approaches, however, inevitably suffer from limitations such as restricted coverage, low efficiency, and high safety risks. With the advancement of unmanned technology, unmanned intelligent terminals are gradually being applied to assist humans in emergency scene sensing. Partheepan \textit{et al.} \cite{Partheepan2023} explored the application of UAVs in forest fire management. The authors of \cite{Zhang2023,wu2023cooperative,Valarmathi2023} investigated the application of unmanned intelligent systems for environmental monitoring and target recognition during natural disasters such as earthquakes and tsunamis. They proposed various heterogeneous-entity collaboration frameworks to improve the efficiency and effectiveness of sensing.

\textit{In this work, we focus on deep collaboration among heterogeneous sensing entities, including human workers, UAVs, and UGVs, with the goal of maximizing the task completion rate within a limited available time. The well-designed collaboration mode is different from those in previous works.}

\subsection{Heterogeneous-entity collaborative sensing}
In addition to static sensor networks and mobile crowdsensing participants, unmanned intelligent terminals such as UAVs and UGVs have increasingly active in MCS, undertaking tasks that are  challenging for humans to complete unaided \cite{Beycimen2023,Lyu2023}. For instance, Zhu \textit{et al.} \cite{Zhu2023} designed a crowd-assisted hybrid sensing framework by integrating dedicated sensing vehicles (DSVs) with crowdsensing participants. Recognizing that UGVs cannot fly and UAVs have limited battery endurance, Wang \textit{et al.} \cite{Wang2023a} proposed an air-ground collaboration model, in which UGVs transport and charge UAVs, travel close to target zones, and then deploy the UAVs for sensing. Building upon this, Zhao \textit{et al.} \cite{Zhao2024} introduced a multi-window and diffusion model-based approach to enhance UGV motion strategies and improve training data quality. Focusing on disaster scenarios, Guan \textit{et al.} \cite{Guan2023b} investigated post-disaster communication restoration in severely affected areas. By utilizing UAVs and UGVs as emergency base stations, they developed a trajectory optimization algorithm based on the Multi-Agent Deep Deterministic Policy Gradient (MADDPG) reinforcement learning method. Han \textit{et al.} \cite{Han2024} accounted for the distinct capabilities of human workers, UAVs, and UGVs in sensing tasks, proposing a QMIX-based MARL algorithm for task allocation in disaster response scenarios. 

\textit{While previous research offers valuable insights, its limitations in environmental modeling, agent collaboration, and task representation  hindering its direct application to HECTA in emergency rescue. Addressing this gap, our work focuses on tackling these challenges by dedicatedly designing agent collaboration policy and modeling operations under partial observability, which distinguishes it from prior approaches.}

\subsection{DRL-based Task Allocation Algorithms}
Deep reinforcement learning (DRL) typically models the task allocation process as a MDP, using deep neural networks estimate action values and optimizing network parameters iteratively for effective task allocation schemes. Recent work leverages DRL and MARL for diverse sensing challenges. For instance, Xu \textit{et al.} \cite{Xu2023} introduced CQDRL, adding communication to decentralized DRL-based task allocation. Wang \textit{et al.} \cite{Wang2024b} proposed DRL-UCS for UAV data collection, balancing data freshness and coverage under energy constraints. 
Several studies address air-ground collaboration. For example, Ye \textit{et al.} \cite{Ye2023} developed the h/i-MADRL framework, in which an i-EOI module and an h-CoPO module are designed to enhance both the individuality and cooperation of agents.  
Wang \textit{et al.} \cite{Wang2023a} used the GARL framework with attention-based graph networks to handle road topology for UAV-UGV data collection. Zhao \textit{et al.} \cite{Zhao2024} introduced gMADRL-VCS,   applying diffusion models and hierarchical MARL to optimize navigation and path planning strategies for sensing entities. Furthermore, to address complex task relationships, Zhao \textit{et al.} \cite{Zhao2024a} employed heterogeneous graph RL in HGRL-TA for handling various complex relationships in interdependent multi-task sensing allocation.

\textit{While these studies have made progress in their respective research areas, they do not directly address the three challenges addressed in this paper. We rigorously formulate the problem as a Dec-POMDP and propose a novel MARL algorithm with tailored designs for effective task allocation under emergency rescue scenarios.}

%% file: 3_model.tex
\section{System Model}\label{3}
This section first models the entities involved in emergency rescue scenarios, including the environment, sensing tasks and heterogeneous entities. Subsequently, the sensing process is formulated as an optimization problem, with the objective of maximizing the TCR within a given time constraint. Finally, the problem is modeled as a Dec-POMDP.

\subsection{Modeling of the emergency rescue environment}
\begin{figure}[!t]
\centering
\subfloat[Obstacle modeling]{\includegraphics[width=1.7in]{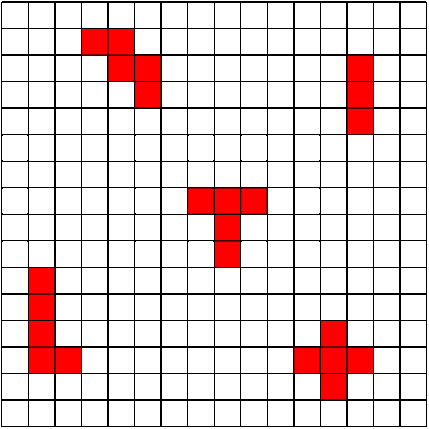}%
\label{障碍物}}
\hfil
\subfloat[Task modeling]{\includegraphics[width=1.7in]{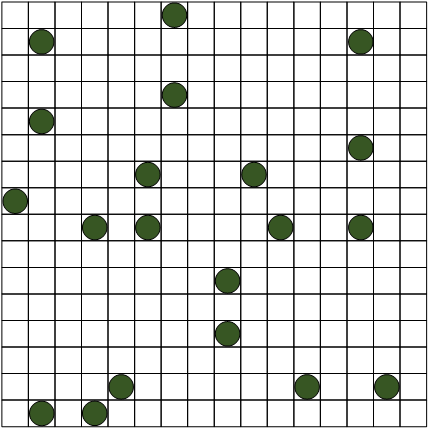}%
\label{任务建模}}
\caption{Grid-based modeling of the emergency rescue environment.}
\label{随机网格}
\end{figure}

To investigate the core issue of task allocation, existing studies \cite{Han2024,Xu2023,Wang2021} often simplify complex environments by abstracting them into 2D or 3D grid representations. Following this convention, this paper employs a grid-based method to model the emergency rescue environment. The sensing area is partitioned into a set of discrete grid cells $P = \{p_1, p_2, \dots, p_m, \dots\}$. Environmental elements, such as obstacles and tasks, are represented within these cells, as illustrated in Fig. \ref{随机网格}.

Each cell $p_m \in P$ can serve as either an obstacle or a task, defined as $p_m=<po_m,pt_m>$. Let $po_m = 1$ if cell $p_m$ contains an obstacle, and $po_m = 0$ otherwise. Similarly, let $pt_m = 1$ if cell $p_m$ corresponds to a potential task location, and $pt_m = 0$ otherwise. We assume a cell cannot be both an obstacle and a task location, as shown in Eq. \ref{公式6}.

The entire sensing process is discretized into $TimeLimit$ time steps, indexed $t \in \{1, 2, \dots, TimeLimit\}$. Each time step represents a sensing period of duration $t_d$. This sensing period serves as the fundamental time unit for decision-making and task allocation. Let $\mathcal{T}_0 = \{\tau_1, \tau_2, \dots, \tau_{N_0}\}$ be the initial set of $N_0$ sensing tasks. Each task $\tau_i \in \mathcal{T}_0$ is defined by a tuple $\tau_i = \langle \tau type_i, \tau loc_i, \tau dur_i\rangle$, representing its type, location, and required execution duration.


\subsection{Modeling of sensing entities}

The heterogeneous sensing workers consist of three types of entities: human workers $\mathcal{W}$, UAVs $\mathcal{D}$, and UGVs $\mathcal{G}$. Let $F = |\mathcal{W}| + |\mathcal{D}| + |\mathcal{G}|$ be the total number of agents. These entities possess distinct characteristics, making them suitable for different types of sensing tasks. In our problem setting, human workers, UAVs, and UGVs operate in a fully cooperative manner, with the proof provided in Appendix \ref{appendix-a}.

At any time step $t$, the state of each agent includes:
\begin{itemize}[leftmargin=*]
    \item \textbf{Human Worker} $w \in \mathcal{W}$: Location $wLoc^t$, movable range $wRge^t$ (set of reachable cells within a time step).
    \item \textbf{UAV} $d \in \mathcal{D}$: Location $dLoc^t$, remaining power $dPow^t$, power consumption per step $dCsp$, movable range $dRge^t$.
    \item \textbf{UGV} $g \in \mathcal{G}$: Location $gLoc^t$, movable range $gRge^t$, UAV detection range $gDet^t$. A UGV $g$ can detect the power of any UAV within its detection range $gDet^t$.
\end{itemize}

We define three types of sensing tasks: high-altitude sensing, fast ground sensing, and detailed sensing which are performed respectively by UAVs, UGVs and human workers. It is assumed that each task type can \textit{only} be completed by the corresponding entity type.

$\textbf{WTRA}$, $\textbf{DTAR}$, and $\textbf{GTRA}$ represent the route sets of different sensing entities. Taking $\textbf{WTRA}$ as an example, its components are the locations of all human workers at each time step, which can be expressed as $\textbf{WTRA} = \{..., wLoc_{k1}^1, ..., wLoc_{k1}^{TimeLimit}, ...\}$, where $k_1$ represents any human worker.

UGVs and UAVs adopt a ``Hard-Cooperative'' battery policy. The ``Hard-Cooperative'' policy means that if an UGV $k_3$ detects that a UAV $k_2$ cannot move due to insufficient power, then at the next time step, the UGV $k_3$ will immediately go to replace battery for the UAV $k_2$ instead of executing the sensing task, as shown in Eq. \ref{公式1}. In this equation, $gLoc_{k_3}^{t+1}$ is the location of the UGV $k_3$ at time step $t+1$, $gDet_{k_3}^{t+1}$ is the range within which the UGV $k_3$ can detect the power of UAVs at time step $t+1$, and $dCsp_{k_2}$ is the power consumption of the UAV $k_2$ during movement.
\begin{align}
\begin{split}
&gLoc_{k_3}^{t+1} = dLoc_{k_2}^t \quad \\
&\textbf{if} \quad dLoc_{k_2}^t \in gDet_{k_3}^t \land dPow_{k_2}^t < dCsp_{k_2}\label{公式1}
\end{split}
\end{align}

For the battery replacement of UAV $k_2$, the following constraints apply: If UGV $k_3$ replace battery for UAV $k_2$, the power of UAV $k_2$ becomes 1. If UGV $k_3$ does not replace battery for UAV $k_2$, and the power of UAV $k_2$ is less than the power required for the next time step movement, the power of UAV $k_2$ remains unchanged. If the power of UAV $k_2$ can support the next time step movement, UAV $k_2$ moves, and the power decreases accordingly. In Eq. \ref{公式2}, $dLoc_{k_2}^t = gLoc_{k_3}^t$ indicates that UAV $k_2$ is at the same location as UGV $k_3$.

\begin{align}
dPow_{k_2}^{t+1} = 
\begin{cases}
1 & \textbf{if } dLoc_{k_2}^t = gLoc_{k_3}^t \\
dPow_{k_2}^t - dCsp_{k_2} & \textbf{if } dLoc_{k_2}^t \neq gLoc_{k_3}^t \\
&\land \ dPow_{k_2}^t \geq dCsp_{k_2} \\
dPow_{k_2}^t & \textbf{if } dLoc_{k_2}^t \neq gLoc_{k_3}^t \\
&\land \, dPow_{k_2}^t < dCsp_{k_2}
\end{cases}
\label{公式2}
\end{align}



\subsection{Optimization problem definition} \label{subsec:optimization_problem}
The objective of the HECTA problem is to determine the optimal sets of trajectories $\textbf{WTRA},\textbf{DTRA},\textbf{GTRA}$ for all agents that maximize the TCR at the final time step $TimeLimit$. Let $N(t)$ be the number of uncompleted tasks at the beginning of time step $t$, with $N(1) = N_0$. The optimization objective is:

\begin{align}
\textbf{max} \frac{ N_0 - N(TimeLimit) }{N_0}\label{公式3}
\end{align}

This optimization is subject to the following constraints for all agents $k \in \mathcal{W} \cup \mathcal{D} \cup \mathcal{G}$, all tasks $\tau_i \in \mathcal{T}_0$, all grid cells $p_m \in P$, and all time steps $t \in \{1, \dots, TimeLimit\}$:

\subsubsection{Obstacle avoidance} Agents cannot occupy grid cells marked as obstacles.
        \begin{align}
        \label{公式4}
        (w/g/d)Loc^t_k.po=0
        \end{align}
    \subsubsection{Movement constraints} Agent movement between consecutive time steps is constrained by their reachable range.
        \begin{align}
        \label{公式5}
        (w/g/d)Loc^{t+1}_k\in (w/g/d)Rge^t_k 
        \end{align}
    \subsubsection{Environment exclusivity} A grid cell cannot simultaneously be an obstacle and a task location.
        \begin{align}
        \label{公式6}
        po_m + pt_m \le 1
        \end{align}
    \subsubsection{Agent task assignment} An agent can be assigned to work on at most one task per time step. Let $x_{ki}^t = 1$ if agent $k$ works on task $\tau_i$ at time $t$, $x_{ki}^t = 0$ otherwise.    
        \begin{align}
        \label{公式7}
        \sum_{i=1}^{N_0} x_{ki}^t \le 1,k=1,2,...
        \end{align}
    \subsubsection{Task agent assignment} A task can be worked on by at most one agent per time step, assuming non-collaborative execution on the same task instance.
        \begin{align}
        \label{公式8}
        \sum_{k \in \mathcal{W} \cup \mathcal{D} \cup \mathcal{G}} x_{ki}^t \le 1,i=1,2,...
        \end{align}
    \subsubsection{Task completion logic} The sufficient and necessary condition for completing a sensing task is that the continuous stay time of a sensing entity in a task area equals the execution time of that sensing task, as expressed in Eq. \ref{公式9}, where $Tloc_i$ denotes the location of sensing task $\tau_i$.

    \begin{align}
    \begin{split}
            &N(t)=N(t-1)-1 \\
        &\textbf{if}\quad \exists k,\ \forall t\in[t,t+\tau dur_i],(w/d/g)Loc^t_k=\tau loc_i
    \end{split}
    \label{公式9}
    \end{align}

The HECTA problem in emergency rescue scenarios can be formulated as follows: Given a set of discrete regions, a set of sensing entities, and time constraints, determine the route sets for the sensing entities to maximize the task completion rates, as shown in Eq. \ref{公式10}. We define the HECTA problem expressed in Eq. \ref{公式10} as \textbf{Problem 1} and subsequently prove its NP-hardness in Appendix \ref{appendix-b}.

\begin{align}
\begin{array}{lll}
&\textbf{confirm}  & \textbf{WTRA},\textbf{DTRA},\textbf{GTRA} 
\\
&\textbf{max} &\frac{N_0-N(TimeLimit)}{N_0}
\\
&\text{s.t.} & (\ref{公式4}),(\ref{公式5}),(\ref{公式6}),(\ref{公式7}),(\ref{公式8}),(\ref{公式9})
\end{array}
\label{公式10}
\end{align}

\subsection{Modeling \textbf{Problem 1} as a Dec-POMDP}

We model \textbf{Problem 1} as a Dec-POMDP, formally defined as an eight-tuple \(\langle S, F, A, T_r, R, \gamma, O, \Omega \rangle\)\cite{Arcieri2024,Lambrechts2024}, where $S$ is the global state space; $F$ is the number of sensing entities; $A$ is the set of sensing entities' actions; $T_r$ is the transition function between states; $R$ is the reward function; $\gamma$ is the discount factor; $O$ is the local observation space for sensing entities; $\Omega$ is the observation function.

\subsubsection{State space}

In emergency rescue scenarios, HECTA involves complex interactions. The global state must capture the relationship between multiple sensing entities and tasks, interactions among entities, and the spatial configuration of obstacles, tasks, and entities. Therefore, the global state \(s^t \in S\) at time step \(t\) can be expressed by Eq. \ref{公式11}:

\begin{align}
\begin{split}
    &s^t=\{obstDist_t,taskDist_t,agentDist_t,
\\
&taskAttr_t,taskDur_t,taskAgent_t,\}
\end{split}
\label{公式11}
\end{align}

\begin{figure}[!htp]
\centering
\includegraphics[width=3.1in]{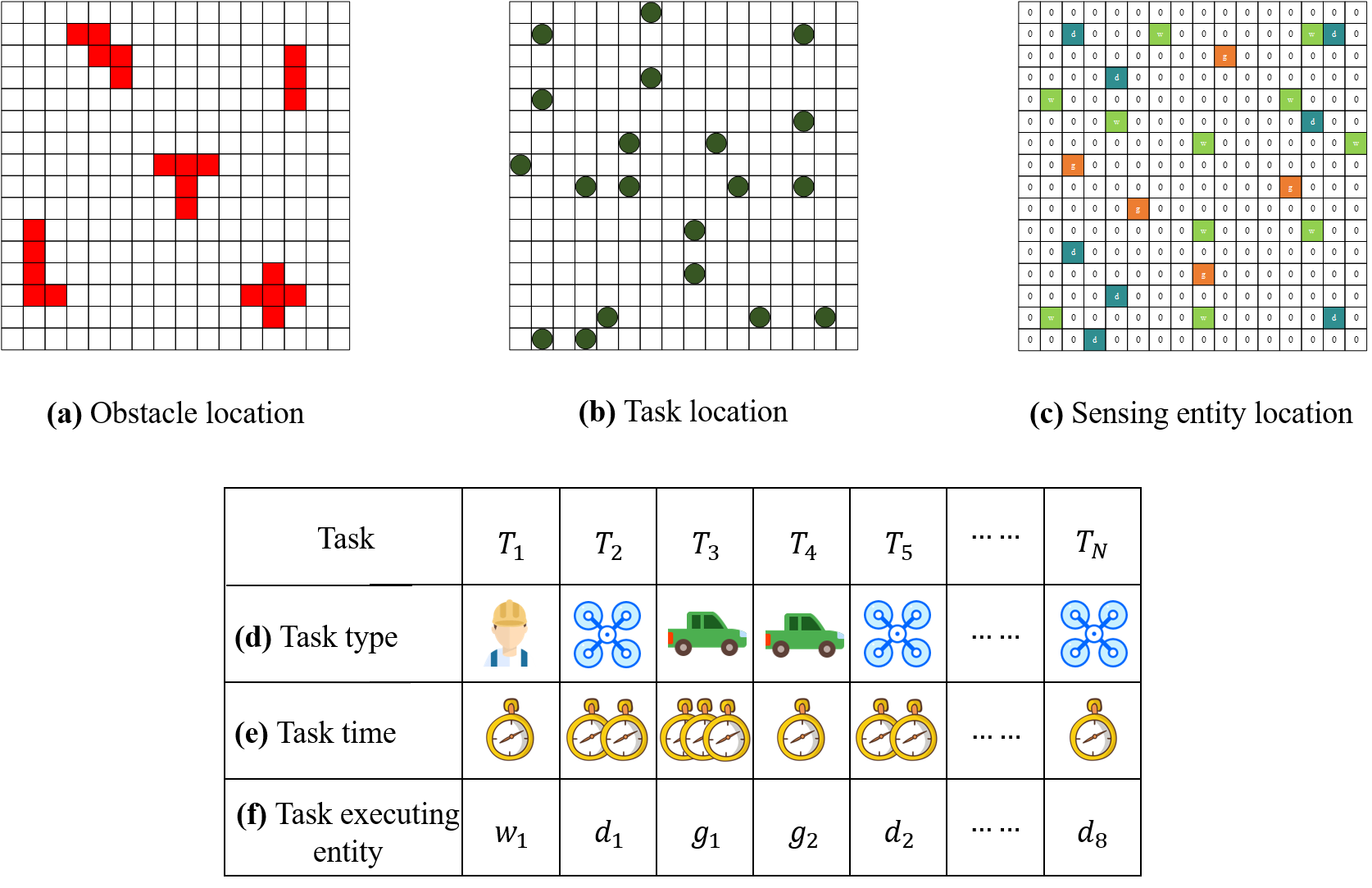}
\caption{State space. (a)-(c) show the distribution of obstacles, tasks and entities respectively; (d) represents the type of task, indicating which category of entity can perform the task; (e) depicts the remaining execution time of the task, which will gradually decrease when an agent is performing the task; (f) shows the matching status between tasks and sensing entities.
}
\label{状态空间}
\end{figure}

Among them, $obstDist_t$ is the representation of static obstacle locations. $taskDist_t$ is the representation of initial task locations. $agentDist_t$ is the representation of sensing entity locations. Each obstacle, task and entity takes a cell \(p_m \in P\) as its location. $taskAttr_t$ is the type of task, indicating which category of entity can perform the task. $taskDur_t$ is the remaining execution time of the task and it gradually decreases when an agent is performing the task. $taskAgent_t$ is the matching status between tasks and sensing entities which indicates whether a task is being executed by an agent and which sensing agent is performing it. Fig. \ref{状态空间} illustrates the state space.

\subsubsection{Local observation space and observation function}

Due to limitations such as communication range and sensor capabilities, each sensing entity \(k\) only has access to partial information about the global state. Thus, the partial observation space of a sensing entity \(k\) at time $t$, denoted as $o_k^t$, is expressed as in Eq. \ref{公式12}. \((w/g/d)Loc^t_k\) is entity's current position. \((w/g/d)Rge^t_k\) is the movable range of entity \(k\) in a time step. \(urge^t_k\) is the power information of an entity, including current power and power consumption per step. Since humans and UGVs do not need to worry about task failure due to energy depletion, we set their \(urge^t_k\) to 0. \(I\!D_k\) is a unique identification for each entity.
\begin{align}
    \begin{split}
        &o^t_k=\{(w/g/d)Loc^t_k, (w/g/d)Rge^t_k,urge^t_k,I\!D_k\}
    \end{split}
    \label{公式12}
\end{align}

\begin{figure}[htbp]
\centering
\includegraphics[width=3in]{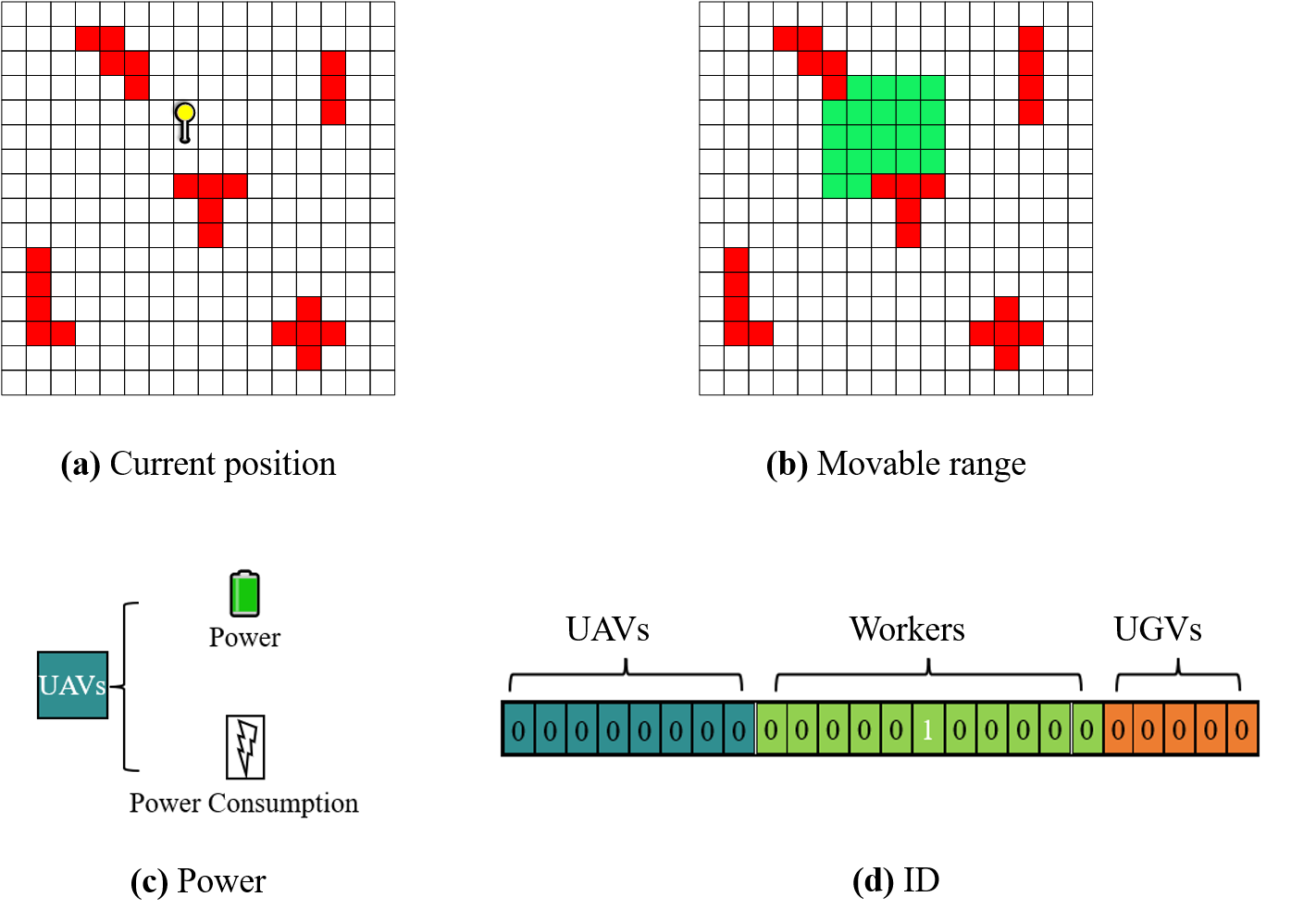}
\caption{Local observation space. (a) represents entity's current position; (b) shows the movable range, the cells highlighted in green, of entity \(k\) within a time step; (c) depicts the power information of UAVs, including current power and power consumption per step; (d) shows entity's unique identification.}
\label{局部观测}
\end{figure}

For ease of computation, one-hot encoding is used for the ID. For the $k$-th sensing entity, the $k$-th bit of its ID is 1, while all other bits are 0. The type of a sensing entity is determined by the position of the bit with a value of 1. Fig. \ref{局部观测} illustrates the partial observation space of sensing entities.

The observation function \(\Omega\) represents the probability of receiving an observation \(o^{t+1}_k\) given the resulting state \(s^{t+1}\) and the action \(a^t_k\) taken, as shown in Eq. \ref{公式观测函数}. This function models the uncertainty and limitations in sensing.

\begin{align}
    \Omega(s^{t+1},a_k^t,o_k^{t+1})=\text{Pr}(o_k^{t+1}|a_k^t,s^{t+1}) \label{公式观测函数}
\end{align}

\subsubsection{Action space}
\begin{figure}[htbp]
\centering
\subfloat[Traditional action space]{\includegraphics[width=1.7in]{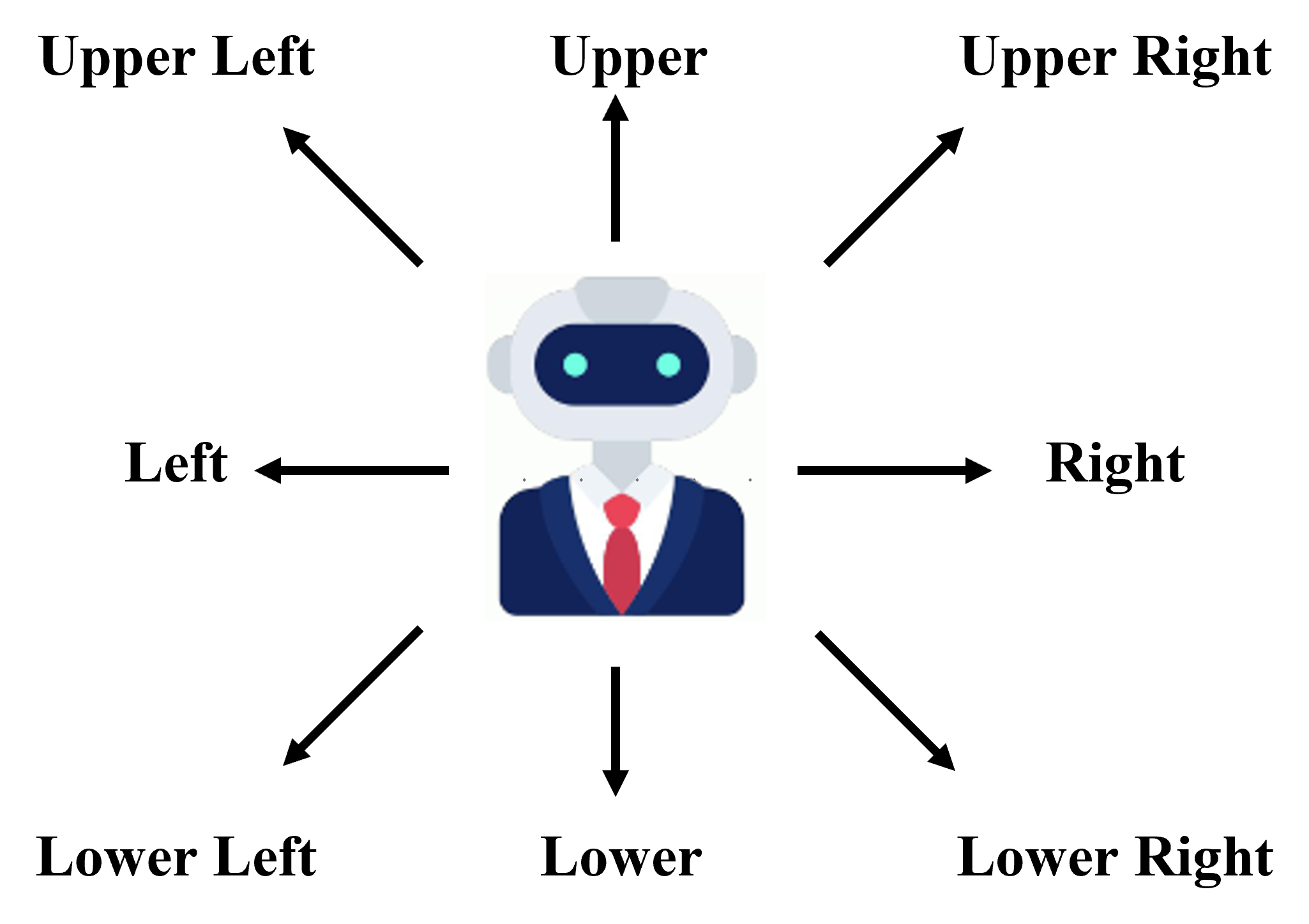}%
\label{传统动作空间}}
\hfil
\subfloat[Action space in this work]{\includegraphics[width=1.4in]{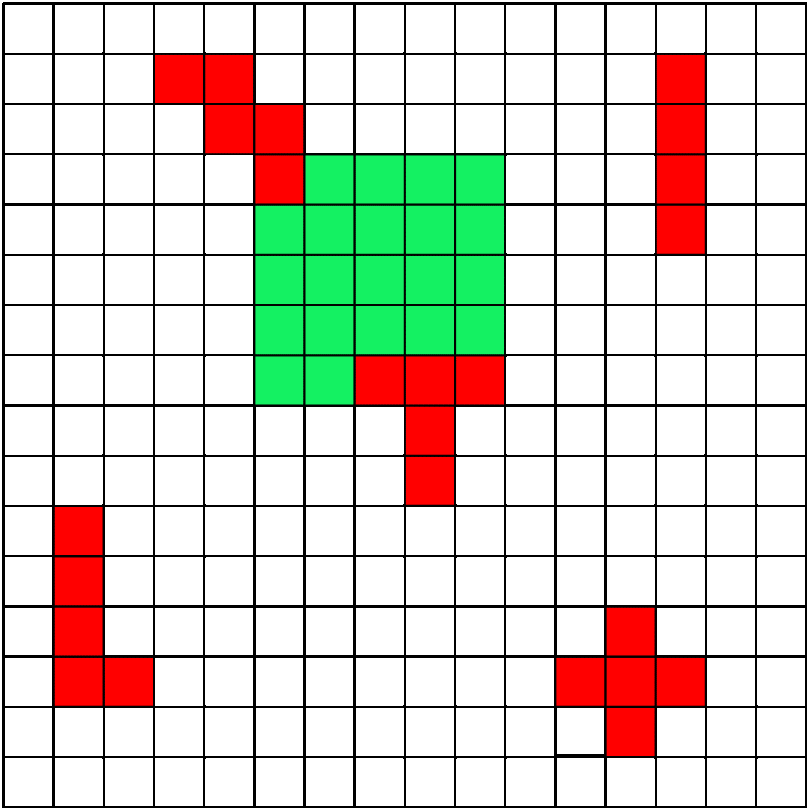}%
\label{特殊动作空间}}
\caption{Action space. (a) shows the traditional action space in route planning problem, which includes discrete movement directions. (b) represents the action space designed in our problem, denoted by the movable range of any entities at the next time step.}
\label{fig_sim}
\end{figure}

In traditional agent route planning problems, the action space often consists of discrete movement directions (e.g., up, down, left, right, stay), as shown in Fig. \ref{传统动作空间}. However, in this paper, we define the potential action space for an agent \(k\) as the set of all possible grid locations \(p_m \in P\) it could intend to move to, as shown in Fig. \ref{特殊动作空间}. Theoretically, sensing entities can reach all cells at a time step. However, due to constraints such as mobility limitations, obstacles, and power, sensing entities can only move to cells within their movable range \(Rge^t_k\) during actual movement, as shown by the green area in Fig. \ref{特殊动作空间}. Therefore, the practical action space of an entity is its movable range, which involves the action filtering mechanism and will be explained in detail later. Thus, the action \(a_k^t\) corresponds to selecting the target location for the next step, as shown in Eq. \ref{公式13}

\begin{align}
    a^t_k = (w/d/g)Loc^{t+1}_k
    \label{公式13}
\end{align}


\subsubsection{State transition and reward function}

The state transition function \(T_r(s^{t+1} | s^t, A^t)\) defines the transition probability from state \(s^t\) to state \(s^{t+1}\) and the joint action can be formulated as \(A^t = \{a_k^t\}^{F}_{k=1}\). 

Given the fully cooperative nature, proved in Theorem 1, all agents share a common reward signal. We define the immediate reward \(r^t\) at time step \(t\) based on the number of tasks completed in that step. Let \(N(t)\) be the number of uncompleted tasks at the beginning of step \(t\). This provides a positive reward only when one or more tasks are completed. We also add a large negative penalty to the reward. If the action selected by a sensing entity exceeds its movable range, $r^{t}$ is set to -10 as a penalty, as shown in Eq. \ref{公式15}.

\begin{align}
r^{t}=\left\{\begin{array}{ll}
N(t-1)-N(t)\ &\textbf{if}\ \ \forall a_{k}^{t} \in(w/g/d)Rge_{k}^{t} \\
\!\;-10\!\;\ &\textbf{else}
\end{array}\right.
\label{公式15}
\end{align}

In practical scenarios, the scene area is much larger than the movement range of sensing entities, making the probability of positive rewards extremely low. This leads to the sparse reward problem, which makes the model difficult to train. Eq. \ref{公式16} calculates the probability of positive rewards at time step \(t\), where $P$ is the set of the 2D region. In practical action selection, adding an action filtering mechanism \cite{Han2024} can effectively solve this problem. Before executing an action, the sensing entities first perform a logical check to determine if the action is within the movement range. If not, they reselect the action.

\allowdisplaybreaks
\begin{align}
pro_t = \prod_{k_1} \frac{wRge_{k_1}^t}{\left| P \right|} \times \prod_{k_2} \frac{dRge_{k_2}^t}{\left| P \right|} \times \prod_{k_3} \frac{gRge_{k_3}^t}{\left| P \right|}
\label{公式16}
\end{align}

\subsection{Belief state}
Define $h^t_k$ as the history information, which includes all action selections and local observations, and represent $h^t_k$ as $h^t_k=\{o^0_k,a^0_k,...,o^{t-1}_k,a^{t-1}_k,o^t_k\}$. Since $h^t_k$ is a variable that is related to time, its length grows linearly, which is not conductive to our storage. Therefore, we define a probability distribution over the state space, namely the belief state $b_k(s^t)$, as shown in Eq. \ref{公式信念空间}, where $b_0$ is the belief state at the beginning. Since the state space is finite, the belief space is also finite. We rigorously prove in Appendix \ref{appendix-c} that the belief state \(b_k(s^t)\) satisfies the Markov property, serves as the foundational assumption for HECTA4ER.

\begin{align}
    b_k(s^t)=P(s^t|h^t_k,b_0)   \label{公式信念空间}
\end{align}

%% file: 4_method.tex
\section{Methodology}\label{4}
\begin{figure*}[!t]
\centering

\includegraphics[width=6.5in]{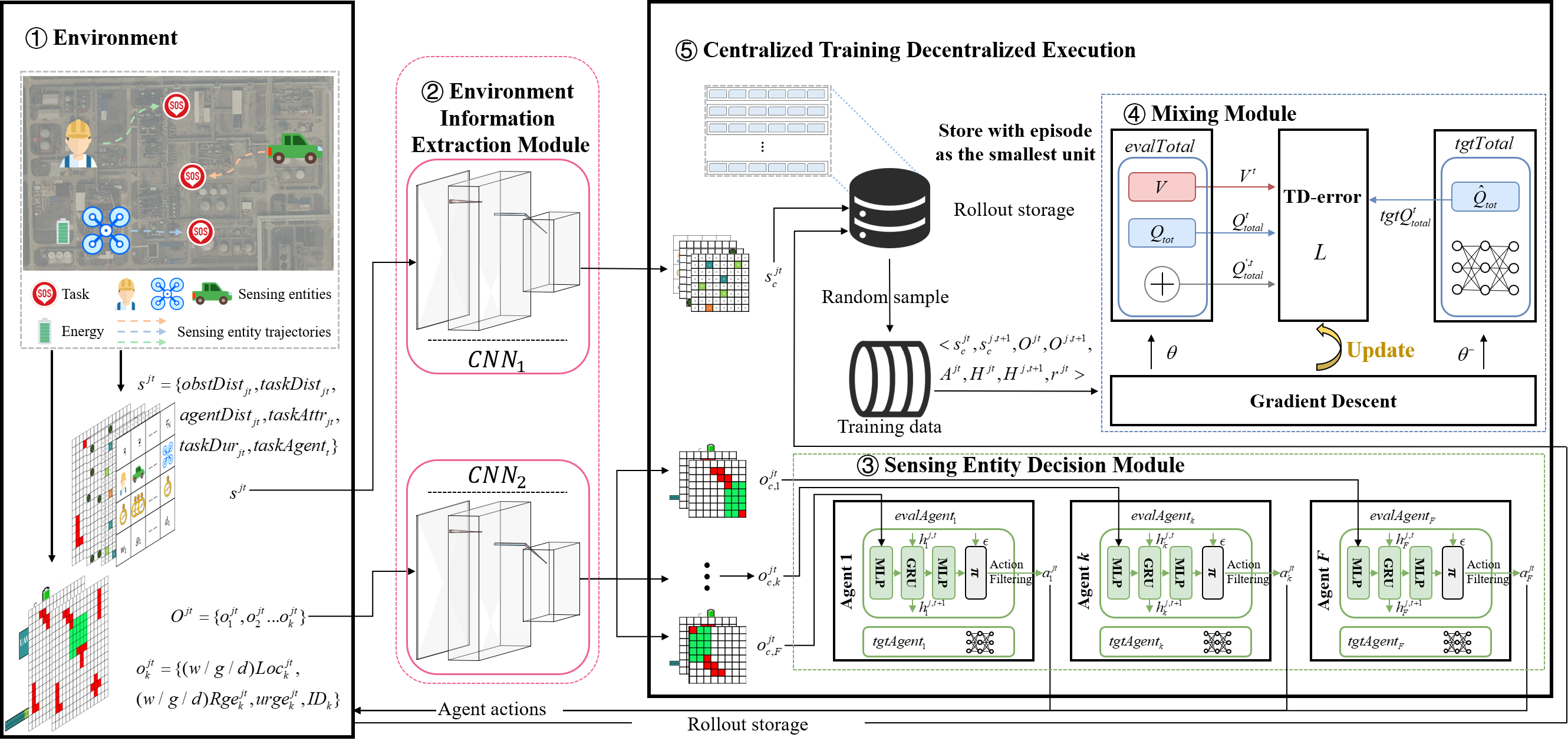}%
\caption{Overview of the HECTA4ER algorithm. The workflow begins with sensing entities interacting with the environment to obtain global states and local observations, which are then processed by the EIEM to extract relevant features. These features are used by the SEDM to make individual decisions, selecting actions within the entities' movable ranges. The Mixing Module integrates these decisions to compute a global value for the system. All interaction experiences are stored in the replay buffer. During training, the algorithm samples from this buffer to update the networks, optimizing the parameters to improve task allocation efficiency. In the execution phase, entities act based on local observations using their individually trained networks, enabling decentralized yet coordinated task execution.}
\label{算法结构图}
\end{figure*}

To address the HECTA problem, this section details the proposed task allocation algorithm called HECTA4ER. The architecture of HECTA4ER is illustrated in Fig. \ref{算法结构图}. Following a description of its constituent modules, a convergence analysis of the algorithm is presented in Appendix \ref{appendix-d}.

\subsection{Environment information extraction module}

Part 1 of Fig. \ref{算法结构图} illustrates the environment with which heterogeneous entities interact. Entities obtain the global state $s^{jt}$ and local observations $O^{jt}$ from the environment and input them to the environment information extraction module (EIEM), part 2 of Fig \ref{算法结构图}. $j$ is the memory index of replay buffer $E$, which will be introduced later. The EIEM consists of two Convolutional Neural Networks (CNNs)\cite{Yuan2023} with identical structures but different parameters. 

The global state $s^{jt}$ and local observations $O^{jt}$ are input to $CNN_{1,2}$ to extract features and finally obtain latent feature representations $s^{jt}_c$, $\{o^{jt}_{c,k}\}^F_{k=1}$, as shown in Eq. \ref{cnns} and Eq. \ref{cnno}. Each $CNN$ typically comprises fully connected layers, convolutional layers, pooling layers and ReLU activation function.

\begin{align}
        s_c^{jt} &= CNN_1(s^{jt}) \label{cnns}\\
        \{o_{c,k}^{jt}\}^F_{k=1} &= CNN_2(\{o^{jt}_k\}^F_{k=1}) \label{cnno}
\end{align}

\subsection{Sensing entity decision module}
Part 3 of Fig. \ref{算法结构图} illustrates the sensing entity decision module (SEDM). Each agent $k$ has an individual decision network, denoted as $evalAgent_k$. Crucially, parameters are shared across agents of the same type to promote learning efficiency and scalability. The core of $evalAgent_k$ is based on a Recurrent Neural Network (RNN)\cite{Perumal2024} and consists of two Multi-Layer Perceptron (MLP) layers, one Gated Recurrent Unit (GRU) layer, an action selection layer and an action filtering mechanism. The $evalAgent_k$ takes the extracted local observation features $\{o^{jt}_{c,k}\}^F_{k=1}$ as input and outputs the action value $\{Q^{jt}_k\}^F_{k=1}$ for each action in the action space, as shown in Eq. \ref{rnn过程}. The two MLP layers serve as the input and output layers of the SEDM for dimensionality mapping respectively. The GRU layer is used to calculate the action value $\{Q^{jt}_k\}^F_{k=1}$ and maintains a hidden state $\{h^{jt}_{k}\}^F_{k=1}$ that encodes the agent's action-observation history. The action selection layer uses the $\varepsilon$-greedy strategy for action selection. To avoid sparse rewards, we use $(w/g/d)Rge^{jt}_k$ to filter actions during actual training. If the selected action $a^{jt}_k$ is not in $(w/g/d)Rge^{jt}_k$, the hidden state is adjusted and the action selection is performed again. The belief state $b_k(s^{jt})$ is the probability distribution of the state \( s^{jt} \) based on the history of actions and local observations. \textbf{Theorem 3} at Appendix \ref{appendix-c} proves that the belief state has the Markov property. Therefore, it is necessary to use the RNN to process historical information, thereby ensuring that the entire learning process is a Markov process in the belief space. The procedure of SEDM is shown in Algorithm \ref{感知主体决策过程}.

\begin{align}
    Q^{jt}_k = evalAgent_k(o_{c,k}^{jt}, h_k^{jt})\label{rnn过程}
\end{align}

\begin{algorithm}[htbp]
\caption{Entity decision process with action filtering }
\label{感知主体决策过程}
\hangindent=4.5em 
        \hangafter=1
\KwIn{Extracted local observation \(o^{jt}_{c,k}\), \\ \hspace{2.8em} The decision module \(evalAgent_k\),  \\ \hspace{2.8em} Moveable range \((w/g/d)Rge^{jt}_k\), \\ \hspace{2.8em} Exploration parameter \(\varepsilon\) }
\KwOut{ Selected action \(a^{jt}_k\), \\ \hspace{3.8em} Action value \(Q^{jt}_k\),\\ \hspace{3.8em} Hidden state \(h^{jt}_k\).}
    Compute all action values according to Eq. \ref{rnn过程}
    
    \While{True}
    {
        Select action \(a^{jt}_k\) based on the \(\varepsilon\)-greedy policy.
    
        \eIf{\(a^{jt}_k \in  (w/g/d)Rge^{jt}_k \)}{
            \Return \(a^{jt}_k\) , \(Q^{jt}_k\) and \(h^{jt}_k\)\;
        }{Set action value of \(a^{jt}_k\) to \(-\infty\)\;}
    }
    
\end{algorithm}

\subsection{Mixing module}
Part 4 of Fig. \ref{算法结构图} depicts the mixing module, which comprises a central network $evalTotal$ and its corresponding target network $tgtTotal$. The central network $evalTotal$, which is responsible for evaluating the global value of the system, consists of a joint action-value network \(Q_{tot}\), a state-value network \(V\), and a summation unit. Both \(Q_{tot}\) and \(V\) receive the state \(s_c^{t}\), all entities' hidden values set \(H^{t} = \{h_k^t\}_{k=1}^F\), and all entities' actions set \(A^{t} = \{a_k^t\}_{k=1}^F\) as inputs, producing outputs \(Q_{tot}^t\) and \(V^t\), respectively, as formulated in Eq. \ref{mixmodule}. The summation unit aggregates the individual action-values \(Q_k^{t}\) from all entities to compute a sum \(Q_{tot}^{\prime ,t}\), detailed in Eq. \ref{sum}. The target network $tgtTotal$, utilized to stabilize the training of \(evalTotal\), incorporates a target joint action-value network \(\hat{Q}_{tot}\) to generate the target value $tgtQ^t_{total}$, as shown in Eq. \ref{tgt}. \(\bar{A}^t = \{\bar{a}_k^t\}_{k=1}^F\) is the set of actions where each \(\bar{a}_k^t\) maximizes the individual \(Q_k^t\). \textbf{Theorem 1} at Appendix \ref{appendix-a} proves that the heterogeneous entities are fully-cooperative, and thus, the relationship between the joint action-value function $Q^{t}_{total}$ and the individual entities' action-value $\{Q^{t}_k\}_{k=1}^F$ satisfies the monotonicity condition.

\begin{align}
    Q^{t}_{total},V^{t} &= evalTotal(s^{t}_c,H^{t},A^{t})\label{mixmodule} \\
    Q^{\prime,t}_{total} &= \sum_{k=1}^F Q_k^{t}\label{sum}\\
    tgtQ^{t+1}_{total}& = tgtTotal(s^{t+1}_c,\{h^{t+1}_k\}_{k=1}^F,\bar{A}^{t+1})\label{tgt}
\end{align}

\subsection{CTDE training process and loss function}
The algorithm employs the CTDE architecture\cite{Amato2024}. During the centralized training phase, the algorithm leverages global information $s^{t}$ and data aggregated from all entities to train the critic components. The components include the EIEM networks, the SEDM entity networks $evalAgent_k$ and the Mixing Module $evalTotal$. In the decentralized execution phase, each agent $k$ makes decisions based solely on its local observation history $o_k^t$ and $h_k^{t-1}$, using its individually trained network $evalAgent_k$.

\begin{figure}[!t]
\centering
\includegraphics[width=3.5in]{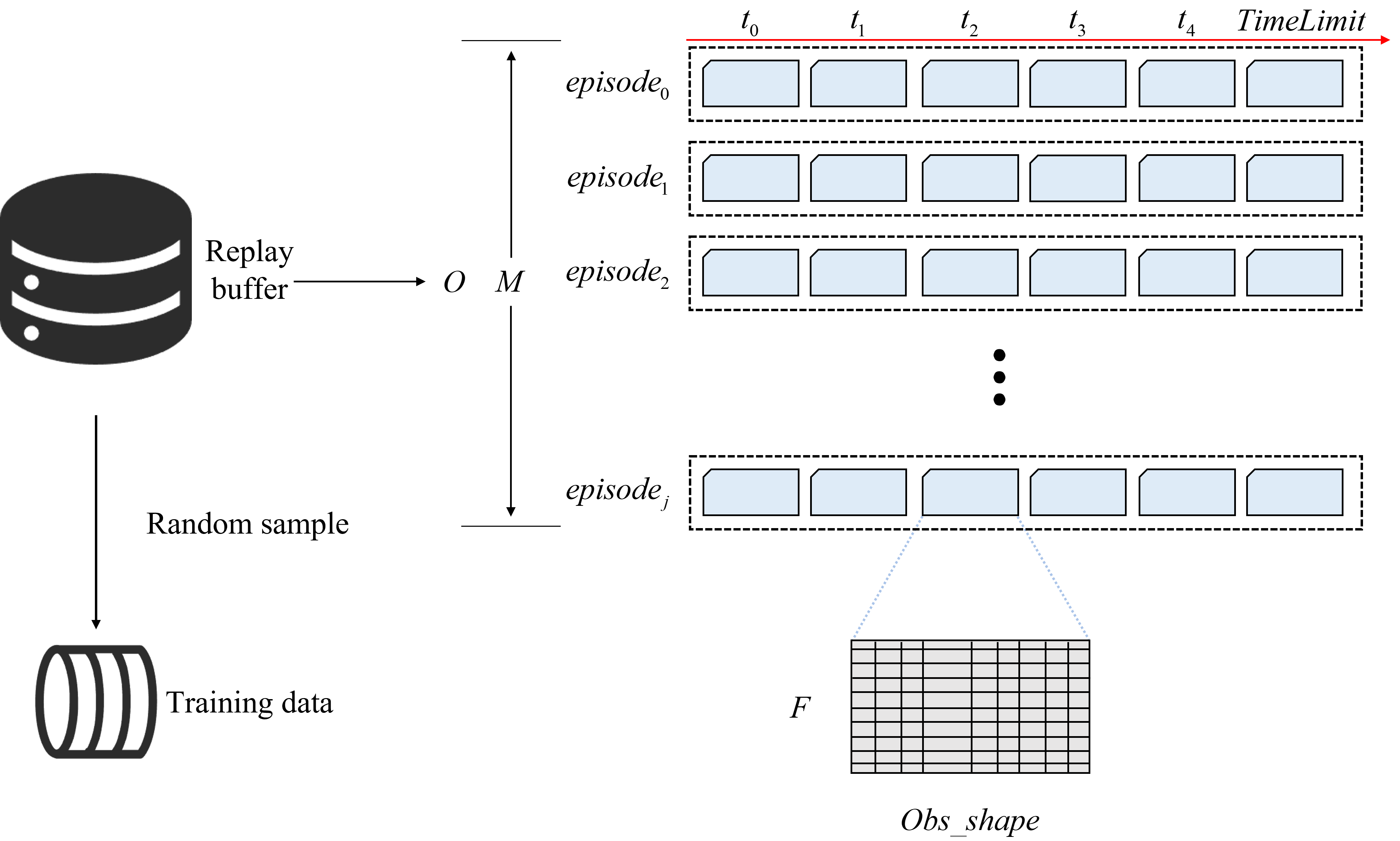}
\caption{Structure of experience replay.}
\label{经验回放}
\end{figure}

Unlike traditional methods that store single-step transitions in the replay buffer, our approach treats a complete episode as the fundamental unit for storage, as depicted in Fig. \ref{经验回放}. Consider the set of local observations \(O^t = \{o_k^t\}_{k=1}^F\) at time step \(t\). The replay buffer, with a capacity of \(M\) episodes, stores sequences of these observation sets along with other relevant transition data (actions, rewards, states, etc.) for multiple episodes, each having a maximum length of \(TimeLimit\) steps. Within each stored episode, the data for each time step includes information from all \(F\) entities.


During training, the algorithm optimizes the entities' policies through simulated interactions with the environment. The procedure involves the following steps: First, the neural networks (\(evalTotal\), \(CNN_{1,2}\) ,\(tgtTotal\), \(evalAgent_k\)) and the replay buffer are initialized. The parameters of the evaluation networks are copied to their corresponding target networks. Second, entities interact with the environment for a maximum of \(TimeLimit\) steps per episode, generating experience trajectories that are stored in the replay buffer. At each time step \(t\) , each entity \(k\) selects an action \(a_k^t\) based on its current observation \(o_k^t\) and hidden state \(h_k^t\). The environment transitions to the next state and provides a reward \(r^t\). Third, after a sufficient number of episodes have been collected, batches of episodes are randomly sampled from the replay buffer. These samples are used to update the parameters \(\theta\) of the evaluation networks (\(evalTotal\) and \(evalAgent_k\)) through gradient descent, minimizing the loss function defined below. Periodically, the updated parameters \(\theta\) of the evaluation networks are copied to the target network parameters \(\theta^-\) to enhance training stability.

The total loss function \(L\) is defined in Eq. \ref{公式19}.  It comprises three components: a temporal-difference (TD) loss \(L_{td}\) and two regularization losses \(L_{opt}\) and \(L_{nopt}\), related to the value decomposition. \(L_{td}\) minimizes the squared difference between the estimated joint action-value \(Q_{tot}^t = Q_{tot}(s^t_c, H^t, A^t; \theta)\) and a target value \(y^{dqn}\). The target \(y^{dqn}\) is calculated using the reward \(r^t\) and the output of the target mixing network \(tgtQ^{t+1}_{total}\) \cite{Huang2024}. Here, \(H^t\) denotes the set of all entities' hidden states at time \(t\), \(\theta\) represents the parameters of the evaluation networks, \(\theta^-\) represents the parameters of the target networks, and \(\gamma\) is the discount factor.

\(L_{opt}\) and \(L_{nopt}\) are consistency losses derived from the value decomposition principle, ensuring that the relationship between the joint value and individual values satisfies the Individual-Global-Max (IGM) condition \cite{Rashid2020}. This condition implies that the optimal joint action corresponds to the combination of optimal individual actions. The mathematical derivation and justification are provided in \cite{Son2021}. In these loss terms, \(\lambda_{opt}\) and \(\lambda_{nopt}\) are weighting coefficients. \(Q^{\prime,t}_{total}\) is the sum of individual Q-values. \(\hat{Q}_{tot}(s^t_c, H^t, A^t)\) represents the output of the target mixing network $tgtTotal$ evaluated at the current state \(s^t_c\). Hidden states \(H^t\), specific actions \(A^t\) and \(V(s^t_c, H^t)\) are the state-value component from the evaluation network $evalTotal$.

\begin{align}
&L=L_{td}+\lambda_{opt}L_{opt}+\lambda_{nopt}L_{nopt}
\nonumber\\
&L_{td}=Q_{tot}(s^t_c,H^t,A^t;\theta)-y^{dqn};\  y^{dqn} = r^t + \gamma tgtQ^{t+1}_{total}
\nonumber\\
&L_{opt}=(Q^{'}_{tot}(s^t_c,H^t,\bar{A}^t)-\hat{Q}_{tot}(s^t_c,H^t,\bar{A}^t;\theta^-)\nonumber\\
&\quad\quad\ \ \ \ \ +V(s^t_c,H^t;\theta))^2
\nonumber\\
&L_{nopt}=(\text{min}[Q^{'}_{tot}(s^t_c,H^t,A^t)-\hat{Q}_{tot}(s^t_c,H^t,A^t; \theta^-)\nonumber\\
&\quad\quad\ \ \ \ \ +V(s^t_c,H^t;\theta),0])^2
\label{公式19}
\end{align}

\begin{algorithm}[!t]
\caption{HECTA4ER}
\label{算法1}
\hangindent=4.5em 
        \hangafter=1
\KwIn{
Environment,  \\ \hspace{3.1em} Episode length \(TimeLimit\),  \\ \hspace{3.1em} Entity number \(F\),  \\ \hspace{3.1em} Target update frequency \(U\),  \\ \hspace{3.1em} Replay buffer size \(M\), \\ \hspace{3.1em} Discount factor \(\gamma\), \\ \hspace{3.1em} Loss weights \(\lambda_{opt}, \lambda_{nopt}\).}
\KwOut{Trained networks parameter \(\theta\).}

    Initialize evaluation network parameters \(\theta\)

    Initialize replay buffer \(E\) with capacity \(M\) episodes
    
    Initialize target network parameters \(\theta^- = \theta\)
    
\For{\(episode=1\ \text{to} \ j\)}
{
    Reset environment and initialize hidden states \(h_k^0\) for all entities \(k\)
    
    \For{timeslot \(t=1\ \text{to} \ TimeLimit\)}
    {
        \For{entity \(k=1 \ \text{to} \  F\)}
        {
            Obtain \(s^{jt}\) and \(o^{jt}_k\)
            
            Obtain \(s_c^{jt}\) and \(o_{c,k}^{jt}\) according to Eq. \ref{cnns} and Eq. \ref{cnno}

            Obtain \(a^{jt}_k\) \(h^{jt}_k\) and \(Q^{jt}_k\) by Algorithm \ref{感知主体决策过程}
            
            Update environment
            
            Obtain \(s^{j,t+1}_c\) and \(o^{j,t+1}_{c,k}\) based on step 9
                    
            Add (\(s_c^{jt}, o_{c,k}^{jt}, a_k^{jt}, o_{c,k}^{j,t+1}, s_c^{j,t+1}, h_k^{jt},Q^{jt}_k)\) to \(E^{jt}_k\)        
        }
        \eIf{\(t==TimeLimit\)}{
            \(te^{jt} = 0\)\;
        }{\(te^{jt} = 1\)\;}
        
        Obtain reward \(r_{jt}\) by Eq. \ref{公式15}
        
        Add \((te^{jt},r^{jt},E^{jt}_k)\) to \(E^{jt}\)
    }
        Add \(E^{jt}\) to \(E^j\)

        \eIf{\(j==M\)}{
            \(j=1\) \;
        }{}
        
        Randomly sample training data from \(E\)

        Obtain \(Q_{tot}, Q^{\prime}_{total},V, tgtQ_{total}\)according to Eq. \ref{mixmodule} - \ref{tgt}
        
        Compute loss \(L\) using Eq. \ref{公式19}
        
        Update \(\theta\) using gradient descent based on \(L\)
        
    \eIf{\(episode \  \% \ U == 0\)}{
        \(\theta^- =\theta\)
    }{}
}
\end{algorithm}

%% file: 5_experiment.tex
\section{Experiments}\label{5}

Extensive experiments are conducted in this section to address the following research questions (RQs):

RQ1: Is HECTA4ER superior to baselines in terms of speed, TCR, and robustness for task allocation involving heterogeneous sensing entities for emergency rescue?

RQ2: What is the performance impact of ablating individual modules?

RQ3: How does algorithm performance change under different scenario parameter settings?

RQ4: To what extent does the algorithm's performance translate to practical scenarios?

The first three questions are explored using simulation data, while the fourth is examined in a real-world setting. This section begins with an overview of the experimental setup, including baseline algorithms, simulation data, hyperparameters, and training procedures. Next, results from the simulation experiments are presented and analyzed in detail. Finally, validation experiments conducted in a real-world scene are conducted to assess the algorithm’s potential and limitations.

\subsection{Experimental settings}

\subsubsection{Baselines} 
We address the HECTA problem in complex emergency rescue scenarios. Baselines for comparison should be able to handle environmental complexity and coordinate these diverse entities.
Therefore, we evaluate our HECTA4ER algorithm against four representative baselines: a greedy algorithm, two DRL methods, and a variant of HECTA4ER. This selection represents key strategies in the field. Crucially, the comparison includes MANF-RL-RP \cite{Han2024}, recognized as the leading contemporary algorithm for this specific task allocation problem, allowing for a robust assessment of HECTA4ER’s capabilities.

\begin{itemize}
    \item \textbf{Greedy-SC-RP:} The greedy strategy computes the Euclidean distance between each sensing entity's movable positions and task locations, selecting the nearest position as the next target.
    \item \textbf{MANF-RL-RP \cite{Han2024}:} A state-of-the-art DRL algorithm for UAV, worker, and UGV collaborative sensing in disaster scenarios, demonstrating superior performance.
    \item \textbf{FD-MAPPO \cite{Yu2022}:} A fully decentralized MARL framework employs independent policy optimization based on PPO. FD-MAPPO is an algorithm capable of coordinating multiple agents without communication dependencies.
    \item \textbf{HECTA4ER-Voluntary:} A ``soft cooperation'' variant of the proposed algorithm. In this version, upon detecting a low-power UAV, the UGV prioritizes its own highest-value action (based on its decision module) rather than being constrained to replenish the UAV's power.
\end{itemize}

Assume the number of sensing entities is \(F\), each sensing entity can choose \(A\) movement positions per time step, and the number of tasks is \(N\). The Greedy-SC-RP algorithm involves distance calculation and position selection, resulting in a total complexity of \(\mathcal{O}(F \times A \times N + F \times A)\). The \(\mathcal{O}(F \times A \times N)\) term corresponds to calculating Euclidean distances between all movable positions and task locations, while \(\mathcal{O}(F \times A)\) term accounts for the position selection process.
The remaining three baseline algorithms and the HECTA4ER algorithm are RL-based and have similar computational complexities. Assuming the total number of neural network parameters is \(B\), the number of training iterations is \(W\), the batch size per iteration is \(batchsize\), and the time length of an episode is \(TimeLimit\), the complexity of these algorithms can be expressed as \(\mathcal{O}(W \times (B + batchsize + TimeLimit))\). Here, the complexity of neural network training is \(\mathcal{O}(W \times (B + batchsize))\), and the complexity of environment state updates and reward calculations is \(\mathcal{O}(W \times TimeLimit)\).

\subsubsection{Data Set}
To train and validate the HECTA4ER algorithm, we first conduct experiments using simulation data. We generated 10 scenarios, each defined by sets of obstacle positions, tasks, and heterogeneous sensing entities, as detailed in Table \ref{仿真数据}. scenario 1-4 differ mainly in entity starting positions to assess their impact on algorithm performance. scenarios 5-10 share the entity positions from scenario 2 but modify other parameters for comparison against it. We based the simulation data on observed patterns in real emergency rescue scenarios to ensure authenticity. Details follow:

\textit{\textbf{Sensing entity:}}
To account for different starting configurations in emergency rescues (from dispatched teams to spontaneously organized groups), we model three initial position distributions for sensing entities: a single shared starting point, a random distribution, and an empirical check-in distribution using real-world data  \cite{Han2023a}. Following references \cite{Zhao2024,Wang2024b,Wang2022}, the ratio of UAVs to workers to UGVs is set at 2:5:1. 
We model distinct mobility capabilities for different entity types, guided by an approximate UAV:human:UGV speed ratio of 8:3:5 from \cite{li2022modeling}. 
To account for potential minor differences within a type, we consider two cases: (1) identical capabilities for all same-type entities, and (2) capabilities randomly generated within a narrow range for each entity. UAV energy consumption is modeled similarly: either uniform across all UAVs or randomly assigned per UAV within a set range.

\textit{\textbf{Obstacle:}}
The number of obstacles impacts the simulation's realism and challenge. Insufficient obstacles fail to represent emergency rescue conditions, whereas excessive obstacles might over-constrain the agents, potentially simplifying the core problem or impeding solutions, which would reduce the study's relevance. Therefore, balancing realism with the need to analyze task allocation, we set obstacle density at 15\% ± 7.5\% of the environment (covering 7.5\% to 22.5\%). Reflecting the unpredictable nature of obstacle placement in actual disasters, their locations are randomly generated.

\textit{\textbf{Sensing task:}}
The spatial distribution of sensing tasks is influenced by both predictable patterns (e.g., from human activity) and unpredictable randomness (e.g., from force majeure events). Therefore, we model task locations using two approaches: an empirical check-in distribution to capture patterns, and a random distribution to represent unpredictability. 
We also explore how the number of different sensing task types, relative to the number of sensing entities, impacts the algorithm performance. Throughout the experiments, we maintain consistency by using exactly three predefined types of sensing tasks.

\begin{table*}[htbp]
\centering
\caption{Simulation data under various scenarios. ``Initial position'' is the distribution of entity initial position. ``Entity number'' denotes the number of UAVs, humans and UGVs in each category. The ``Task number (distribution)'' column specifies the total number of tasks and their distribution patterns, such as random or check-in distributions. ``Task type number'' indicates the variety of sensing tasks for UAVs, humans and UGVs respectively. ``Task execution time'' lists the execution time requirements for different tasks and their number. The ``Obstacle number'' column shows the number of obstacles in the environment, and ``Movement radius'' refers to the movable range of different entities. ``Power consumption'' represents the energy consumption per step for UAVs. Some scenes have diverse values for certain fields, which is to prepare for robustness experiments.}
\renewcommand{\arraystretch}{2.2} 
\resizebox{1.0\linewidth}{!}{
\begin{tabular}{cccccccccc}
\hline
\hline
Sce. No. & Initial position & Entity number & Environment size & \parbox{1.5cm}{\centering Task number (distribution)} & \parbox{1.5cm}{\centering Task type number} & \parbox{1.8cm}{\centering Task execution time (number)} & Obstacle number & Movement radius & Power consumption \\
\cmidrule{1-10}
1 & Same & 6/15/3 & 16*16 & 120 (Random) & 30/75/15 & 1/2 (96/24) & 20 & 8/3/5 & 0.3 \\

2 & Random & 6/15/3 & 16*16 & 120 (Random) & 30/75/15 & 1/2 (96/24) & 20 & [7,9], [2,4], [4,6] & [0.2,0.4] \\

3 & Check-in & 6/15/3 & 16*16 & 120 (Random) & 30/75/15 & 1/2 (96/24) & 20 &  [7,9], [2,4], [4,6] & [0.2,0.4] \\

4 & Check-in & 6/15/3 & 16*16 & 120 (Check-in) & 30/75/15 & 1/2 (96/24) & 20 & [7,9], [2,4], [4,6] &  [0.2,0.4] \\

5 & Random & \parbox{1cm}{\centering 4/10/2, 6/15/3, 8/20/4} & 16*16 & 120 (Random) & 30/75/15 & 1/2 (96/24) & 20 &  [7,9], [2,4], [4,6] & [0.2,0.4]\\

6 & Random & 6/15/3 & \parbox{1cm}{\centering 12*12, 16*16, 20*20} & 120 (Random) & 30/75/15 & 1/2 (96/24) & 20 & [7,9], [2,4], [4,6] &[0.2,0.4] \\

7 & Random & 6/15/3 & 16*16 & \parbox{1.5cm}{\centering 96/120/144 (Random)} & \parbox{1cm}{\centering 24/60/12, 30/75/15, 36/90/18} & 1/2 (96/24) & 20 &  [7,9], [2,4], [4,6] &  [0.2,0.4] \\

8 & Random & 6/15/3 & 16*16 & 120 (Random) & 30/75/15 & \parbox{2cm}{\centering 1/2 (72/48),\quad \quad  1/2 (96/24),\quad \quad  1/2/3 (72/36/12)} & 20 & [7,9], [2,4], [4,6] &  [0.2,0.4] \\

9 & Random & 6/15/3 & 16*16 & 120 (Random) & \parbox{1cm}{\centering 40/40/40, 30/75/15} & 1/2 (96/24) & 20 & [7,9], [2,4], [4,6] & [0.2,0.4] \\

10 & Random & 6/15/3 & 16*16 & 120 (Random) & 30/75/15 & 1/2 (96/24) & \parbox{1cm}{\centering 20/40/60} & [7,9], [2,4], [4,6] &[0.2,0.4] \\
\hline
\hline
\label{仿真数据}
\end{tabular}
}
\end{table*}

\subsubsection{Hyperparameters and training process}

\begin{figure}[!t]
\centering
\includegraphics[width=3in]{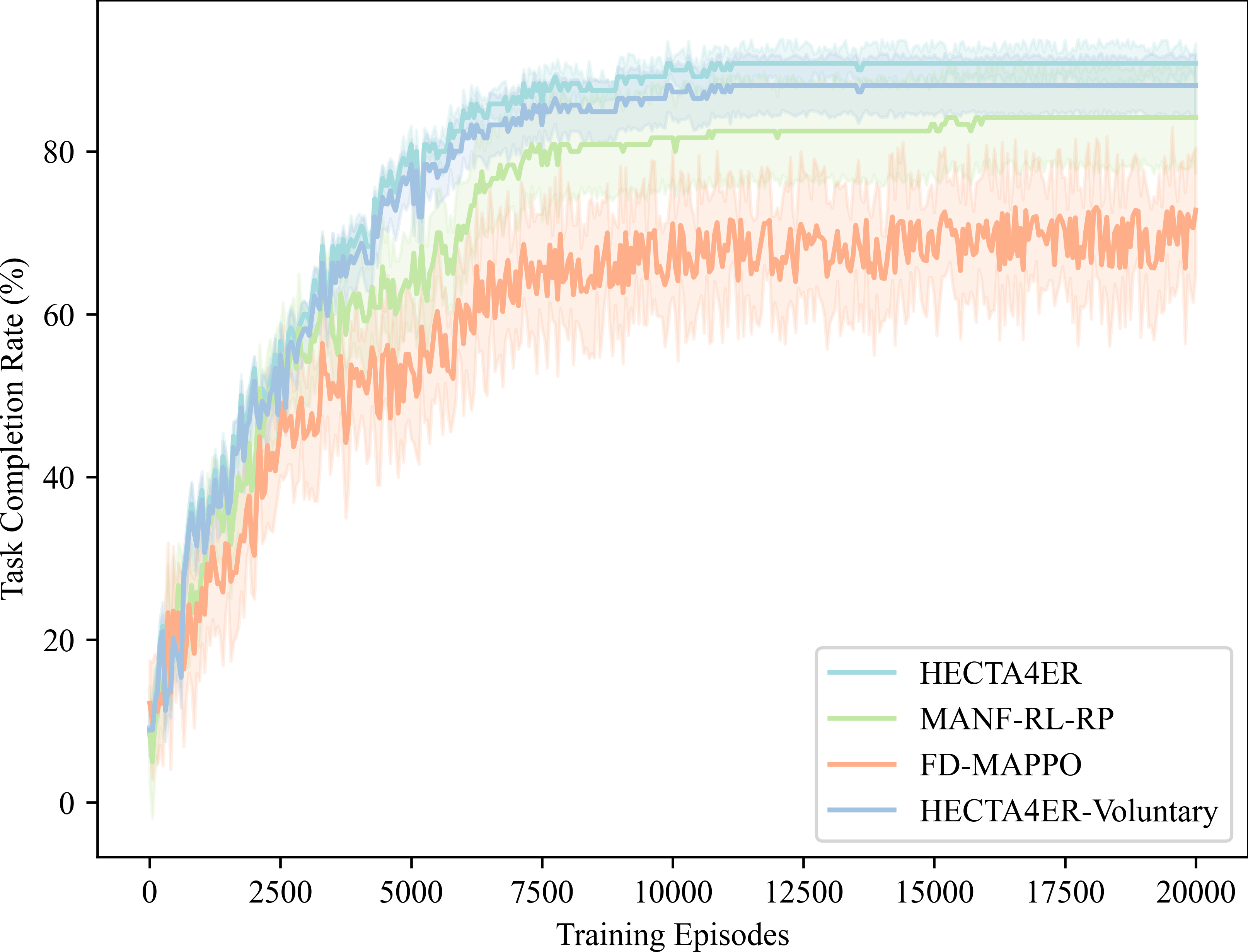}%
\caption{The training curves of different algorithms.}
\label{reward训练效果图}
\end{figure}

The key hyperparameters for HECTA4ER training are set as follows. A common, minimal network architecture is employed across experiments, as network design itself is not the study's focus.  Both RNN and mixing networks have 128 hidden units. The CNN uses 7 input channels (global state), 3 (local observations), 10 output channels, a kernel size of 3, pooling size of 2, and no padding. The replay buffer capacity is 5000, with a clipping coefficient of 0.2. A discount factor of 0.7 prioritizes immediate task execution. We used an RMSprop optimizer with a batch size of 32. The learning rate starts at 1e-4, decaying by 10\% every 1000 episodes to balance exploration/exploitation and ensure convergence (meeting Robbins-Monro conditions). Training and experiments are conducted using simulation data on a system equipped with an Intel Xeon Gold 6430 CPU, an Nvidia RTX 4090 GPU (24GB VRAM), and 120GB of RAM. The software environment comprises PyTorch 2.0.0, Python 3.8 (running on Ubuntu 20.04), and CUDA 11.8. Figure \ref{reward训练效果图} shows the training curves for HECTA4ER and baseline algorithms in scenario 3 ($TimeLimit=12$).
Within the first 6000 training episodes, all algorithms achieve a high TCR. Although MANF-RL-RP converges faster initially, HECTA4ER ultimately reaches a higher final TCR. HECTA4ER-Voluntary performs similarly to HECTA4ER, while FD-MAPPO's TCR is significantly lower than the others.

\subsection{Comparisons with baseline methods (RQ1)}

This section compares the performance of the proposed algorithm with baseline methods in terms of TCR and robustness across different sensing time limits.

\begin{figure}[!t]
\centering
\subfloat[Scenario 1]{\includegraphics[width=1.7in]{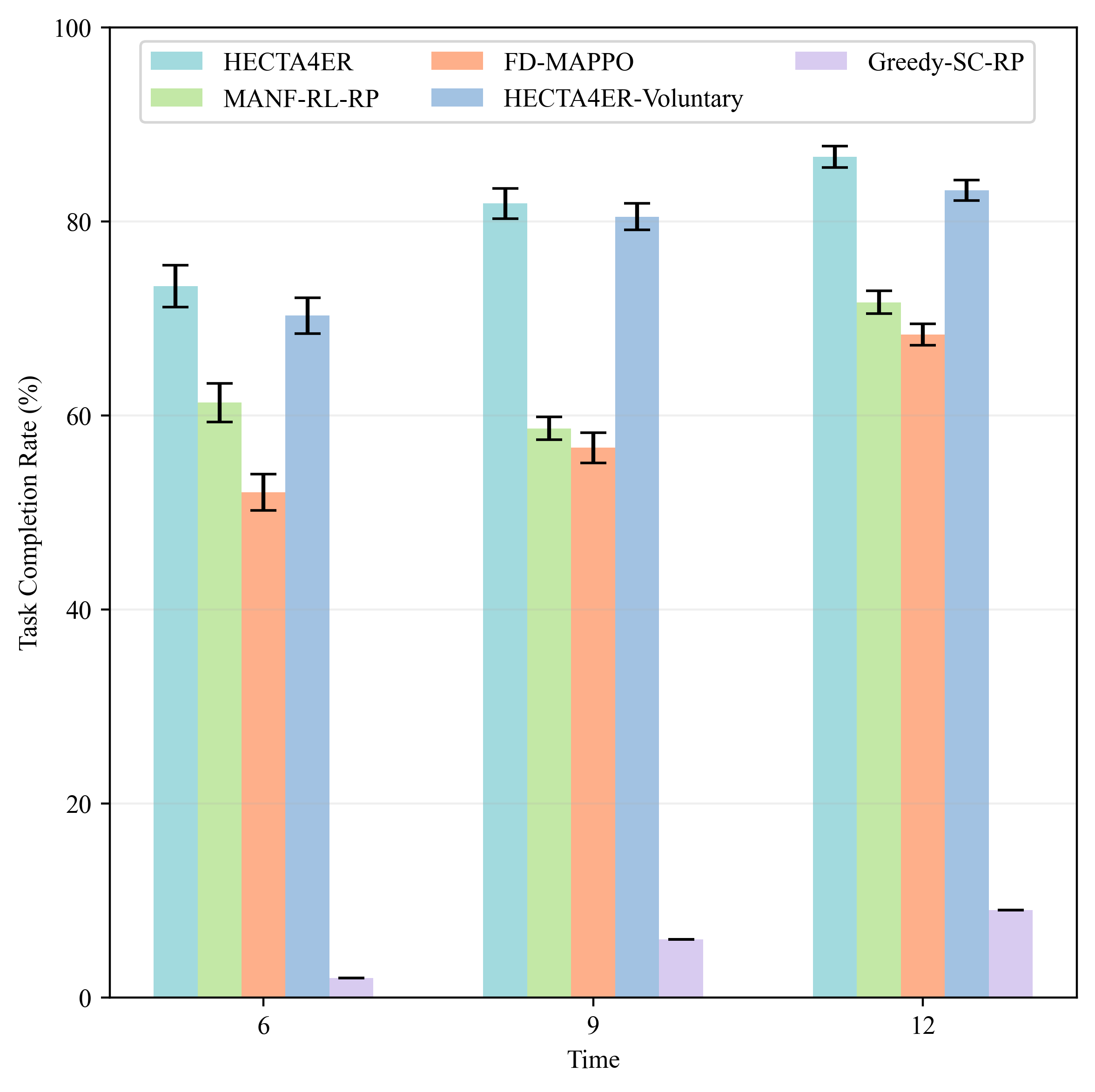}%
\label{Scenario 1}}
\hfil
\subfloat[Scenario 2]{\includegraphics[width=1.7in]{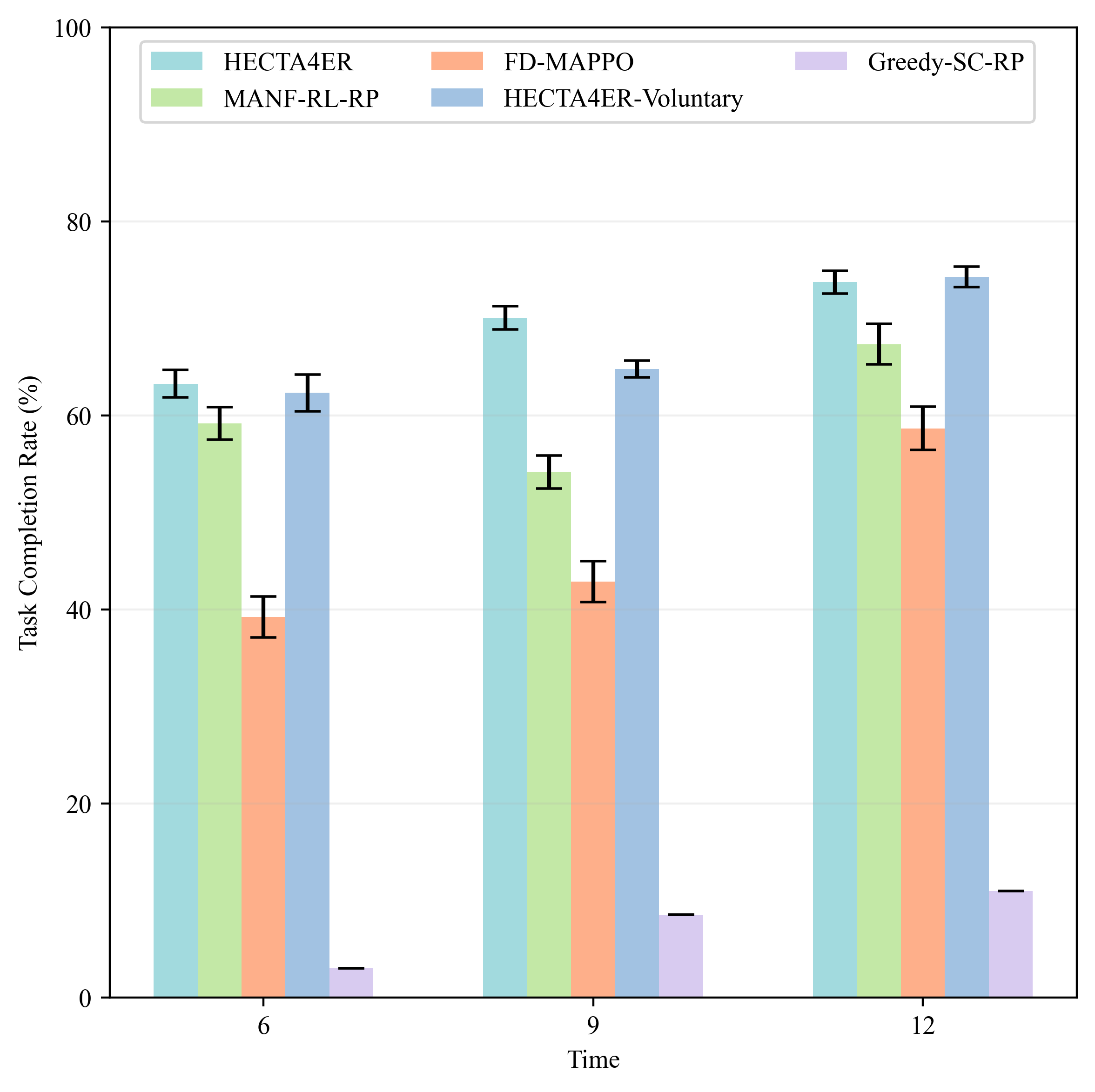}
\label{Scenario 2}}

\subfloat[Scenario 3]{\includegraphics[width=1.7in]{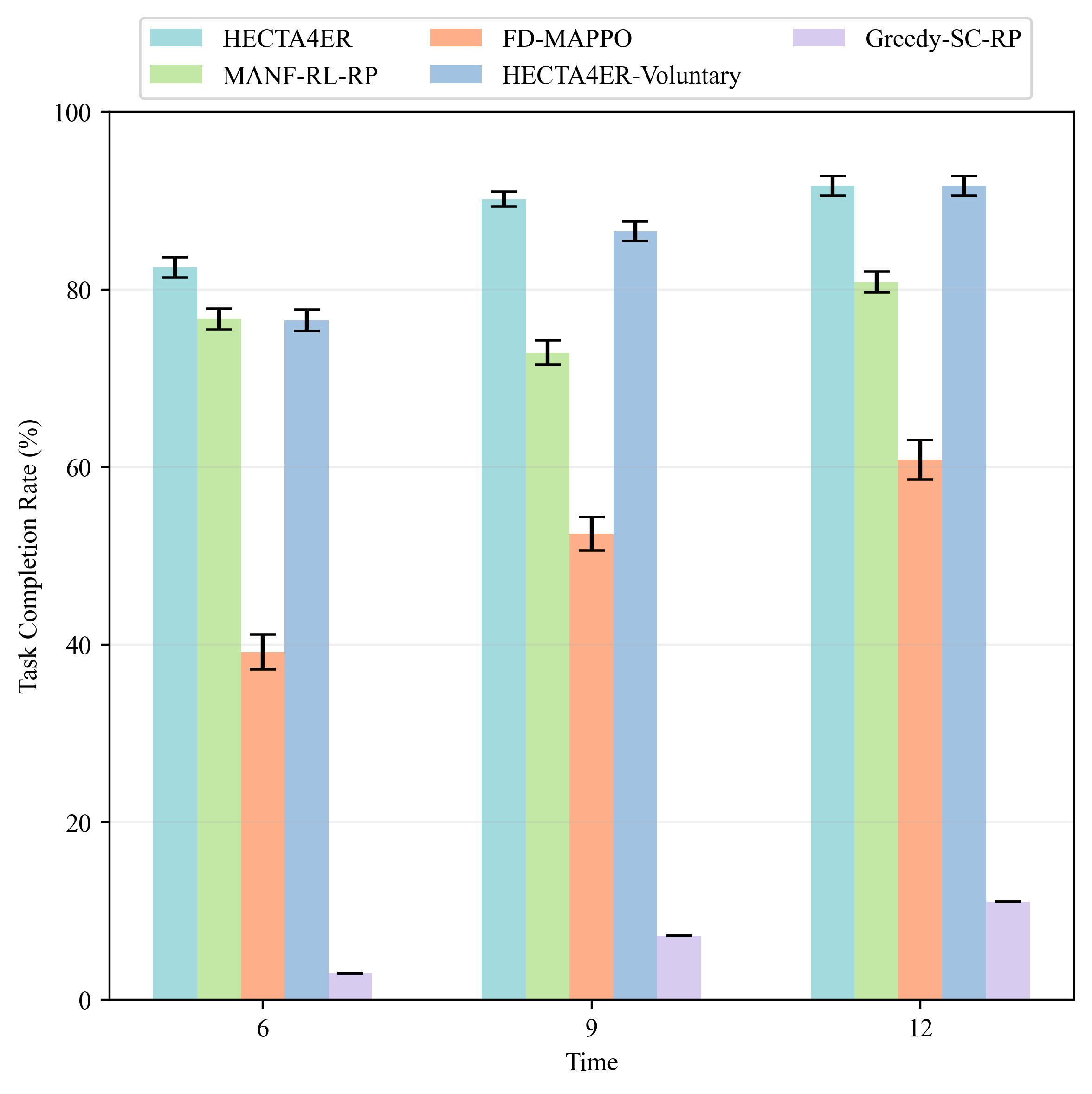}%
\label{Scenario 3}}
\hfil
\subfloat[Scenario 4]{\includegraphics[width=1.7in]{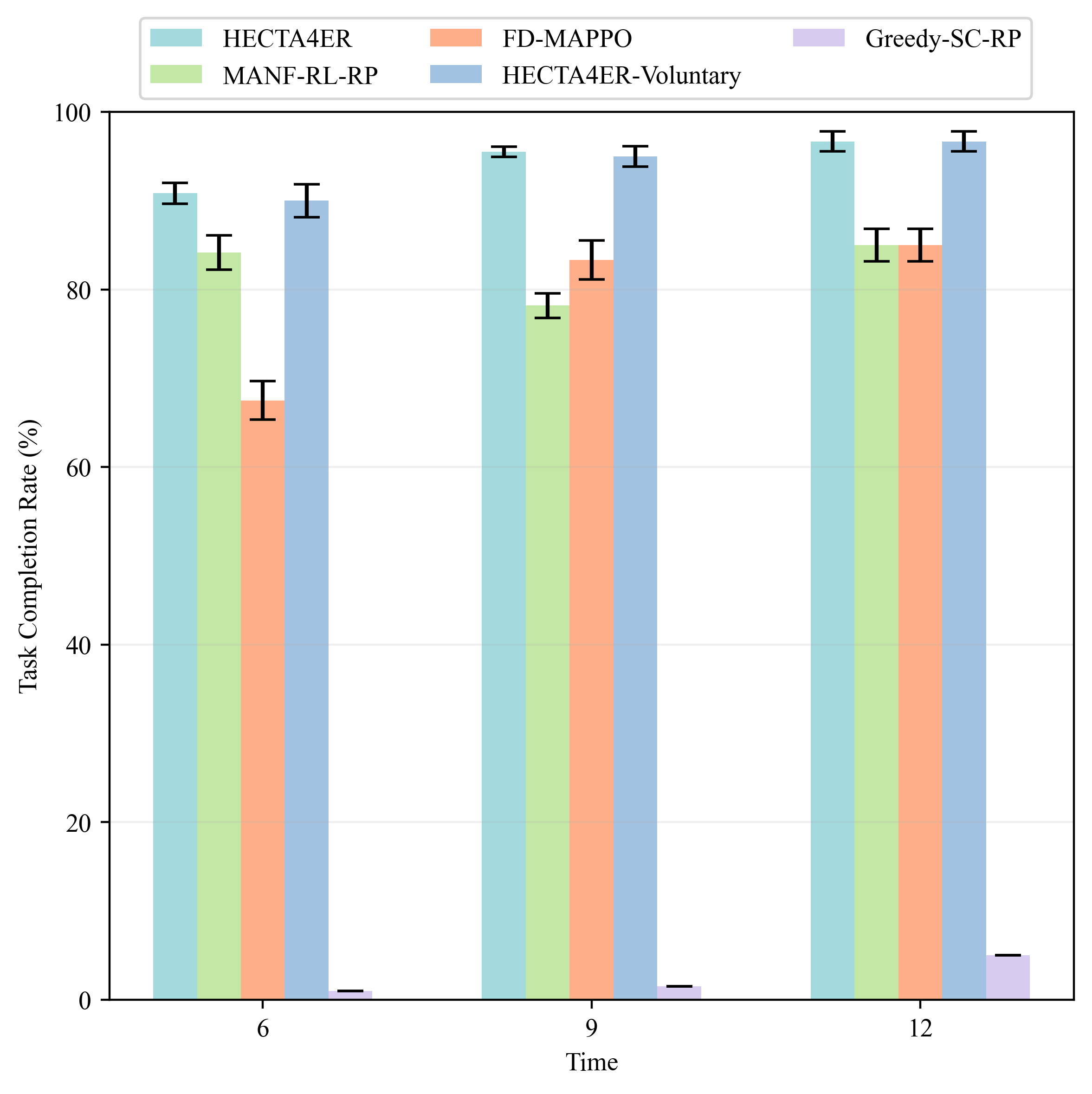}%
\label{Scenario 4}}
\caption{Overall performance of different algorithms with scenario 1-4.}
\label{算法性能对比}
\end{figure}

Figure \ref{算法性能对比} summarizes the results from ten experimental runs (using different random seeds) across scenarios 1-4, with sensing time limits set to 6, 9, or 12 depending on the scenario size. The columns represent the average TCR, and the black bars show the 95\% confidence intervals. The data clearly indicates that extending the sensing time improves TCR by allowing more tasks to be accomplished.

Across scenarios 1 to 4, HECTA4ER consistently achieves the highest TCR. Notably, in scenario 4 (with a sensing time limit of 12), it successfully completes nearly all tasks. The performance of MANF-RL-RP and FD-MAPPO is similar to each other, but significantly lower than HECTA4ER's. HECTA4ER-Voluntary closely matches HECTA4ER's results in most scenarios, though it slightly under-performs in some instances. Greedy-SC-RP consistently yields the lowest TCR among all tested algorithms. The narrow confidence intervals for HECTA4ER and HECTA4ER-Voluntary across scenarios indicate high stability and suitability for reliability-sensitive applications. Conversely, the wider intervals for MANF-RL-RP and FD-MAPPO reflect greater performance variability.

HECTA4ER and MANF-RL-RP are highly comparable as both target the task allocation of heterogeneous entities for emergency rescue. We therefore selected these two algorithms for robustness testing in scenarios 5-10. For robustness testing, the algorithm was trained on a base scenario. Its performance (TCR) was then evaluated across 50 new random scenarios, each generated as a controlled variation of the base scenario. The results are presented in Table \ref{鲁棒性实验结果}.

\begin{table*}[htbp]
\centering
\caption{Experimental results of algorithm robustness comparison  (TCR, \%). The first part of Sce. No. indicates the base scenario, while the second part specifies the variation number. These variations systematically alter parameters such as task execution time, task types, obstacle positions, and entity starting positions to test the algorithm's robustness and adaptability under different conditions. For example, scenario 5-1 represents the first variation of base scenario 5, where entity number is changed to 4/10/2 from 6/15/3.}
\renewcommand{\arraystretch}{1.5}
\resizebox{1.\linewidth}{!}{
\begin{tabular}{ccccccccccc}
\hline
\hline
 \multirow{2}{*}{Sce. No.} & \multicolumn{2}{c}{Training performance} & \multicolumn{2}{c}{Changing task execution time} & \multicolumn{2}{c}{Changing task type} & \multicolumn{2}{c}{Changing obstacle position} & \multicolumn{2}{c}{Changing sensing entity position} \\
\cmidrule(lr){2-3} \cmidrule(lr){4-5} \cmidrule(lr){6-7} \cmidrule(lr){8-9}  \cmidrule(lr){10-11} 
 & MANF-RL-RP & HECTA4ER & MANF-RL-RP & HECTA4ER & MANF-RL-RP & HECTA4ER & MANF-RL-RP & HECTA4ER & MANF-RL-RP & HECTA4ER \\
\hline
5-1 & 51.3 & 72.2 & 32.5 & 44.7 & 18.4 & 28.4 & 56.7 & 67.7 & 18.9 & 30.8 \\

5-2 & 63.2 & 81.0 & 41.5 & 56.1 & 25.9 & 33.2 & 66.3 & 75.1 & 25.6 & 35.9 \\

5-3 & 66.3 & 84.4 & 52.6 & 64.3 & 29.1 & 36.8 & 70.4 & 78.9 & 21.6 & 35.8 \\

6-1 & 63.1 & 86.3 & 46.2 & 63.5 & 28.4 & 36.5 & 62.4 & 77.3 & 43.2 & 43.0 \\

6-2 & 62.5 & 85.0 & 40.9 & 57.0 & 25.1 & 33.6 & 61.9 & 77.8 & 22.6 & 32.2 \\

6-3 & 61.7 & 81.3 & 43.2 & 55.3 & 19.0 & 31.0 & 60.4 & 76.3 & 21.0 & 29.3 \\

7-1 & 60.9 & 84.0 & 46.6 & 61.0 & 24.3 & 35.6 & 68.8 & 76.6 & 25.6 & 36.0 \\

7-2 & 52.5 & 77.0 & 47.3 & 56.0 & 22.5 & 31.5 & 61.5 & 71.9 & 29.8 & 32.7 \\

7-3 & 63.2 & 81.3 & 31.5 & 45.6 & 16.3 & 29.1 & 67.3 & 71.7 & 30.0 & 31.9 \\

8-1 & 49.8 & 62.3 & 32.6 & 43.6 & 15.4 & 21.6 & 50.1 & 57.8 & 24.3 & 30.3 \\

8-2 & 69.0 & 79.0 & 41.6 & 51.0 & 13.6 & 28.0 & 72.1 & 75.3 & 30.5 & 30.4 \\

9-1 & 39.6 & 64.3 & 30.9 & 46.2 & 20.9 & 26.8 & 50.5 & 59.9 & 21.5 & 30.3 \\

9-2 & 62.8 & 82.8 & 43.2 & 55.6 & 24.2 & 30.9 & 65.4 & 76.7 & 25.6 & 31.2 \\

9-3 & 53.6 & 71.0 & 28.1 & 36.2 & 21.6 & 29.9 & 61.5 & 68.7 & 29.4 & 33.8 \\

10-1 & 69.4 & 83.2 & 47.3 & 58.9 & 33.5 & 35.8 & 72.4 & 78.7 & 25.7 & 34.2 \\

10-2 & 70.1 & 78.4 & 48.0 & 55.5 & 25.9 & 31.3 & 63.5 & 71.9 & 24.1 & 32.9 \\

10-3 & 52.4 & 73.2 & 46.3 & 53.1 & 21.3 & 29.0 & 54.3 & 64.6 & 20.9 & 31.8 \\
\hline
\hline
\end{tabular}
\label{鲁棒性实验结果}
}
\end{table*}

Overall, while both MANF-RL-RP and HECTA4ER show performance declines in the controlled random scenarios, HECTA4ER consistently outperforms MANF-RL-RP in each case. This demonstrates HECTA4ER's superior robustness and ability to maintain better performance under environmental uncertainty. Specifically, both algorithms are sensitive to changes in task type and initial sensing entity positions, resulting in the largest performance drops. In contrast, they are more robust to changes in task execution time and obstacle position. This sensitivity stems from task type and initial position changes requiring more complex decision adjustments and environmental reassessment than the less disruptive changes to task duration or obstacles.

\subsection{Ablation study (RQ2)}

We performed ablation studies to evaluate the individual contributions of the EIEM and SEDM modules within HECTA4ER. Using a sensing time limit of 9, ten experiments were conducted for each of scenarios 1-4, with average results presented in Table \ref{消融实验表格}.

\begin{table}[htbp]
\centering
\caption{Experimental results of ablation study (TCR, \%).}
\renewcommand{\arraystretch}{1.5}
\begin{tabular}{lcccc}
\hline
\hline
Algorithm & Sce. 1 & Sce. 2 & Sce. 3 & Sce. 4 \\
\hline
\centering
HECTA4ER                & 82.50 & 70.40 & 89.17 & 95.00 \\
HECTA4ER  w/o EIEM        & 79.17 & 68.37 & 85.83 & 90.00 \\
HECTA4ER  w/o SEDM        & 41.67 & 65.31 & 60.83 & 49.16 \\
HECTA4ER  w/o EIEM \& SEDM & 30.83 & 59.18 & 53.33 & 50.83 \\
\hline
\hline
\label{消融实验表格}
\end{tabular}
\end{table}

The results confirm that both modules play significant roles. Removing the EIEM leads to moderate TCR decreases of 3.33\%, 2.03\%, 3.34\%, and 5.00\% across the four scenarios, respectively. The most notable drop (5.00\% in scenario 4) highlights the EIEM's importance for modeling spatial information. Removing the SEDM results in more severe performance degradation, with TCR reductions of 40.83\% (scenario 1), 28.34\% (scenario 3), and 45.84\% (scenario 4). This underscores the critical role of temporal modeling, handled by the SEDM, especially in dynamic environments. Finally, removing both modules simultaneously causes a substantial 62.7\% drop in TCR for scenario 1 compared to the full model, demonstrating a synergistic effect where EIEM and SEDM together significantly enhance feature representation.

\subsection{Performance comparisons under different scenario parameters (RQ3)}

\begin{figure}[!t]
\centering
\subfloat[Impact of entity number]{\includegraphics[width=1.7in]{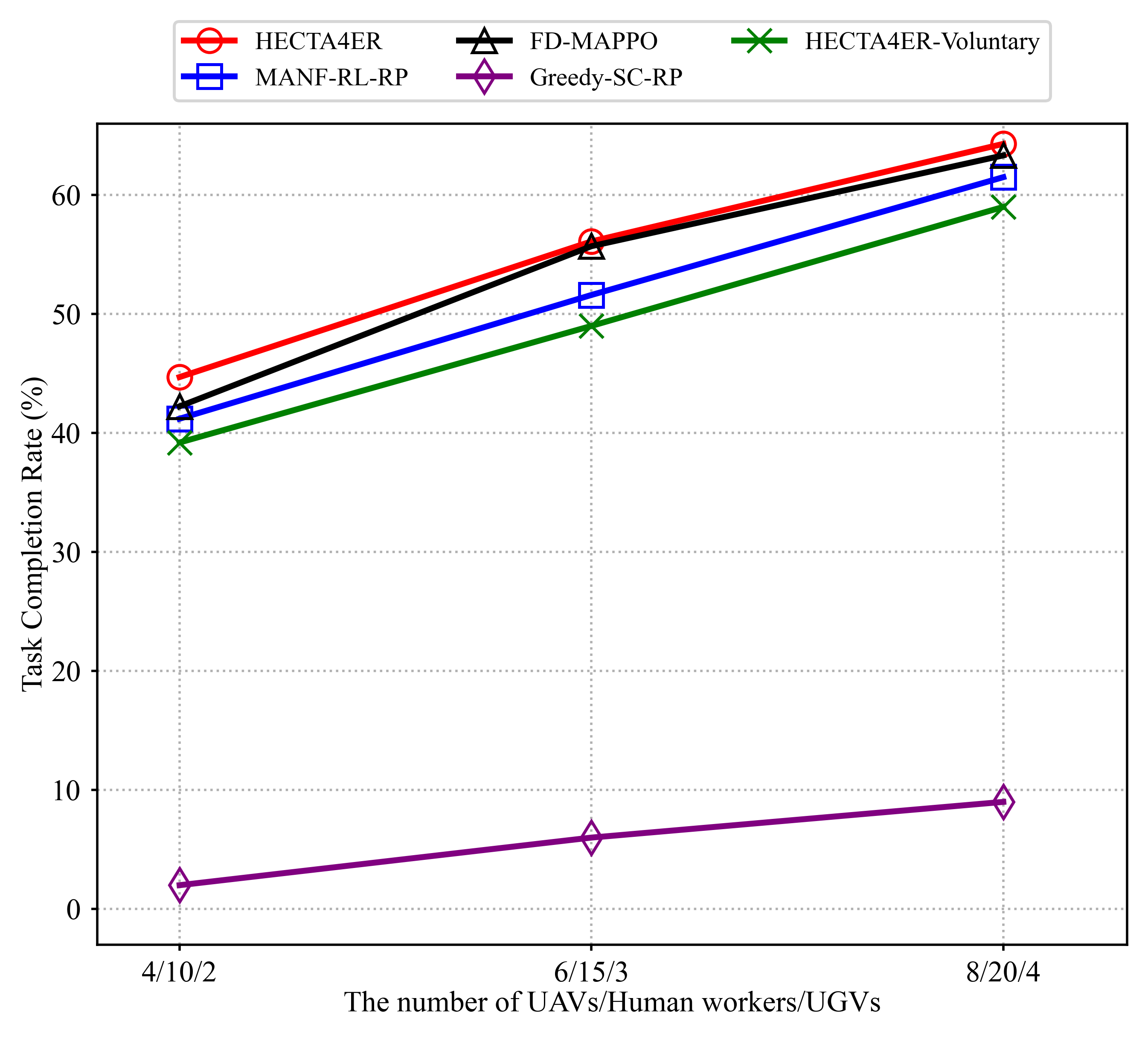}%
\label{Scenario 5}}
\hfil
\subfloat[Impact of environment size]{\includegraphics[width=1.7in]{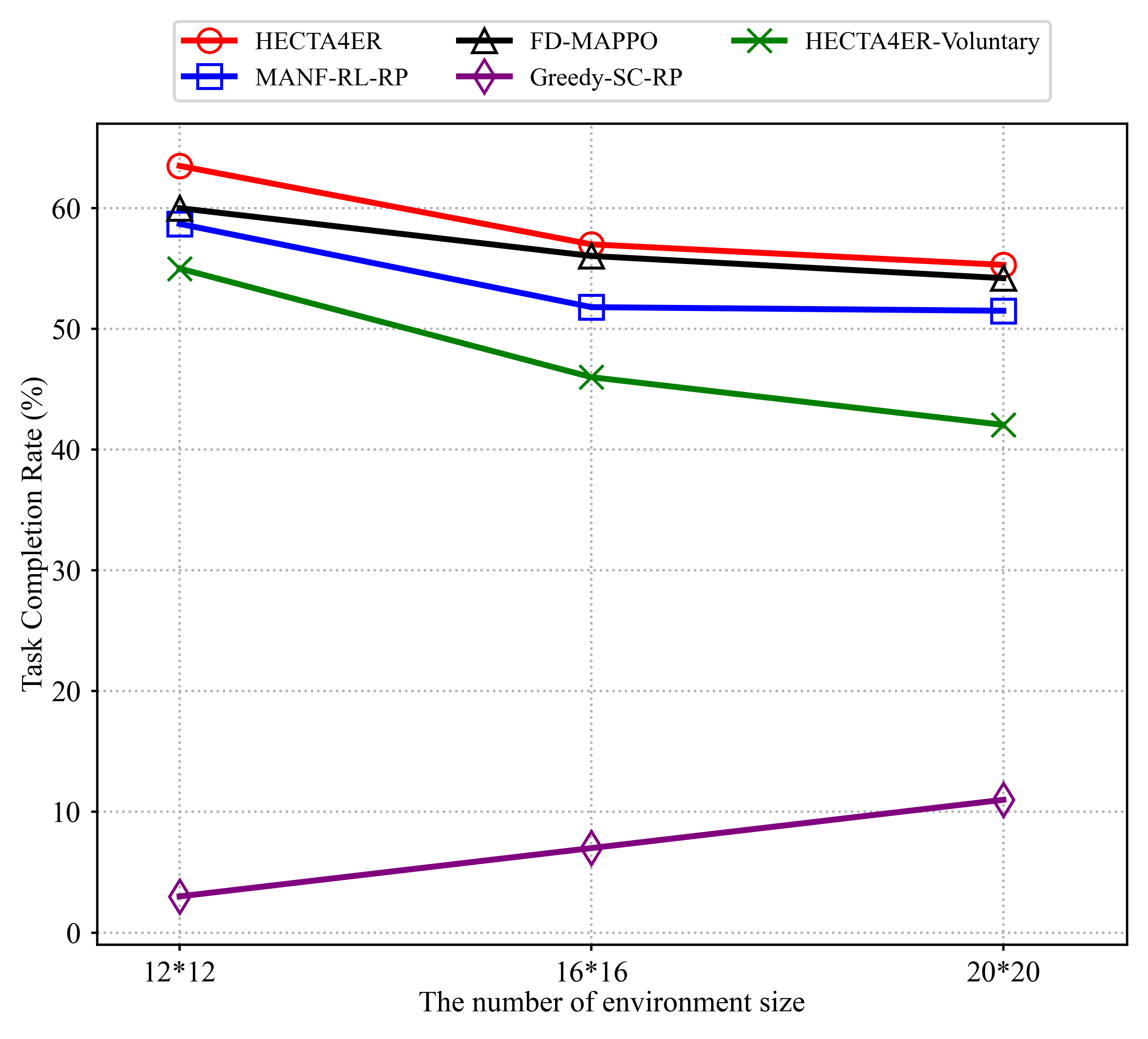}%
\label{Scenario 6}}

\subfloat[Impact of task number]{\includegraphics[width=1.7in]{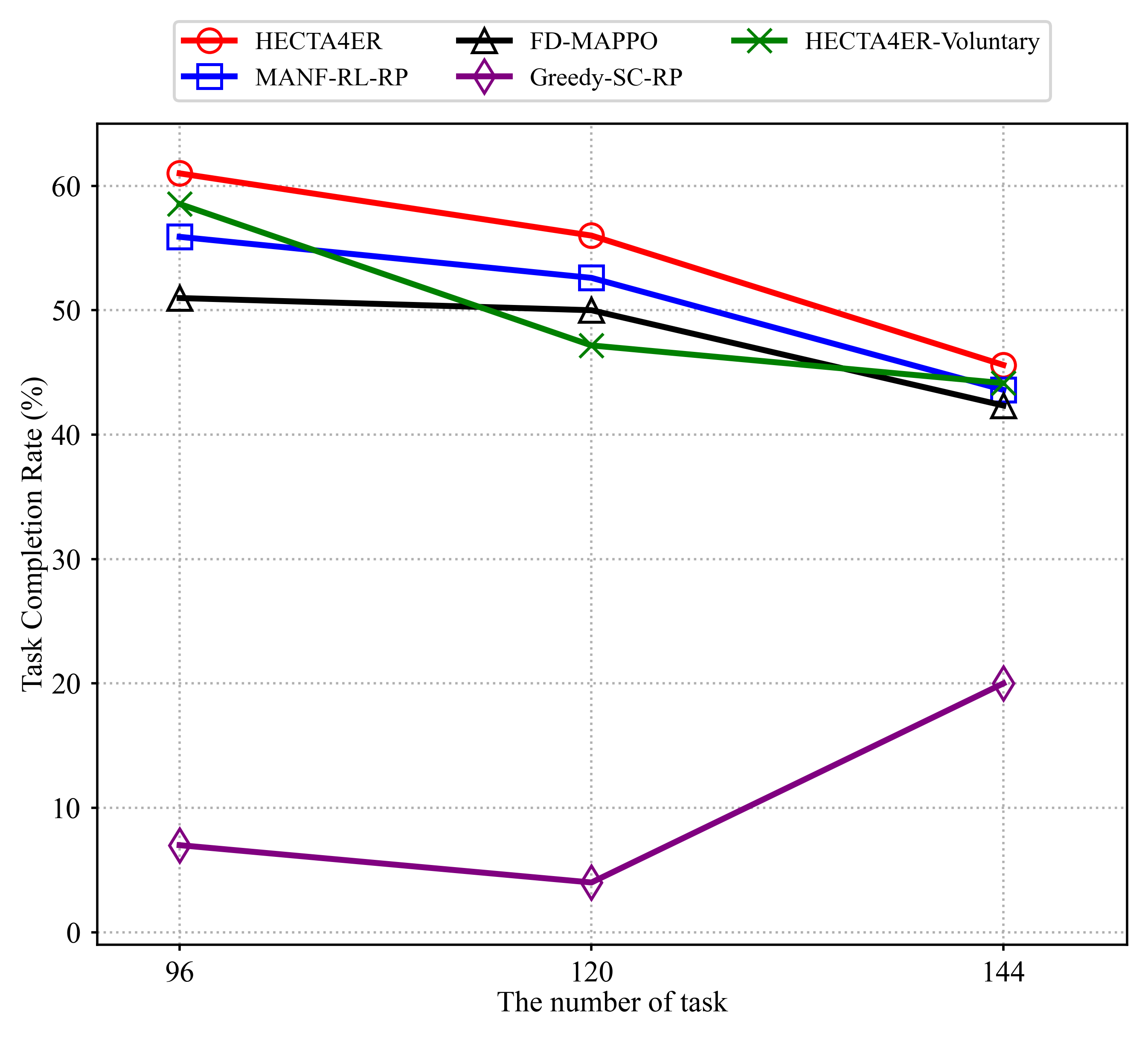}%
\label{Scenario 7}}
\hfil
\subfloat[Impact of task type number]{\includegraphics[width=1.7in]{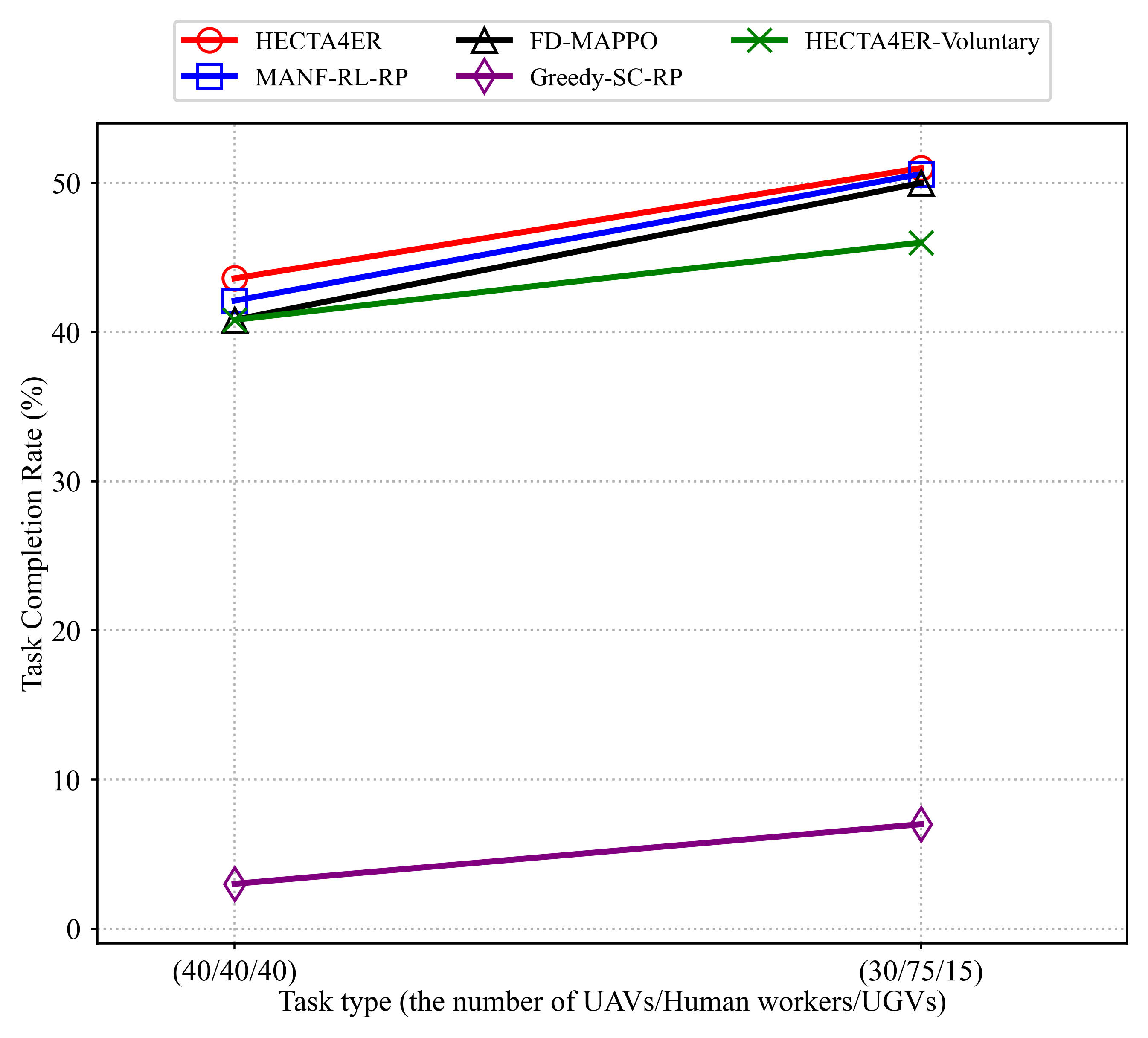}%
\label{Scenario 8}}

\subfloat[Impact of task execution time]{\includegraphics[width=1.7in]{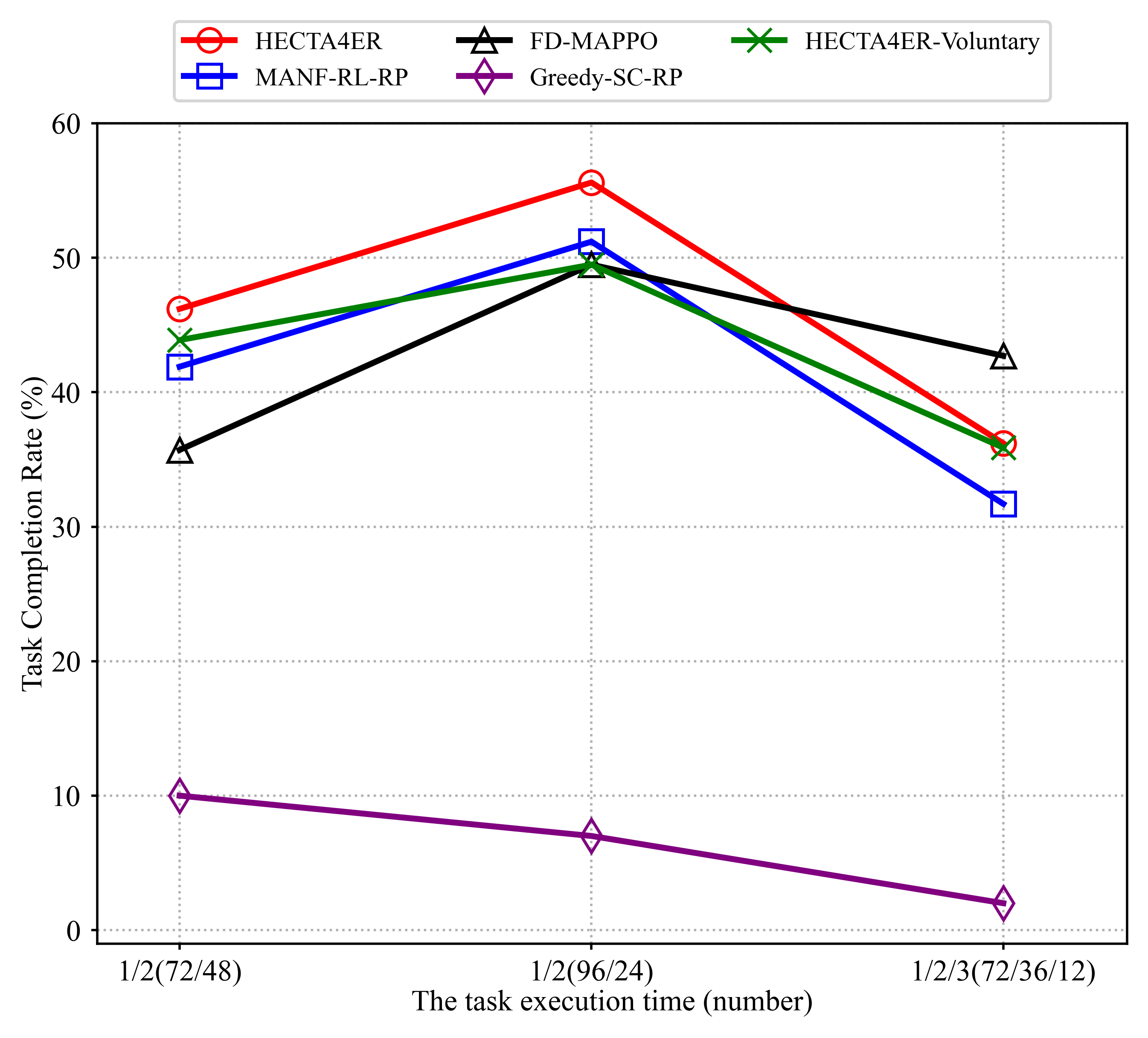}%
\label{Scenario 9}}
\hfil
\subfloat[Impact of obstacle number]{\includegraphics[width=1.7in]{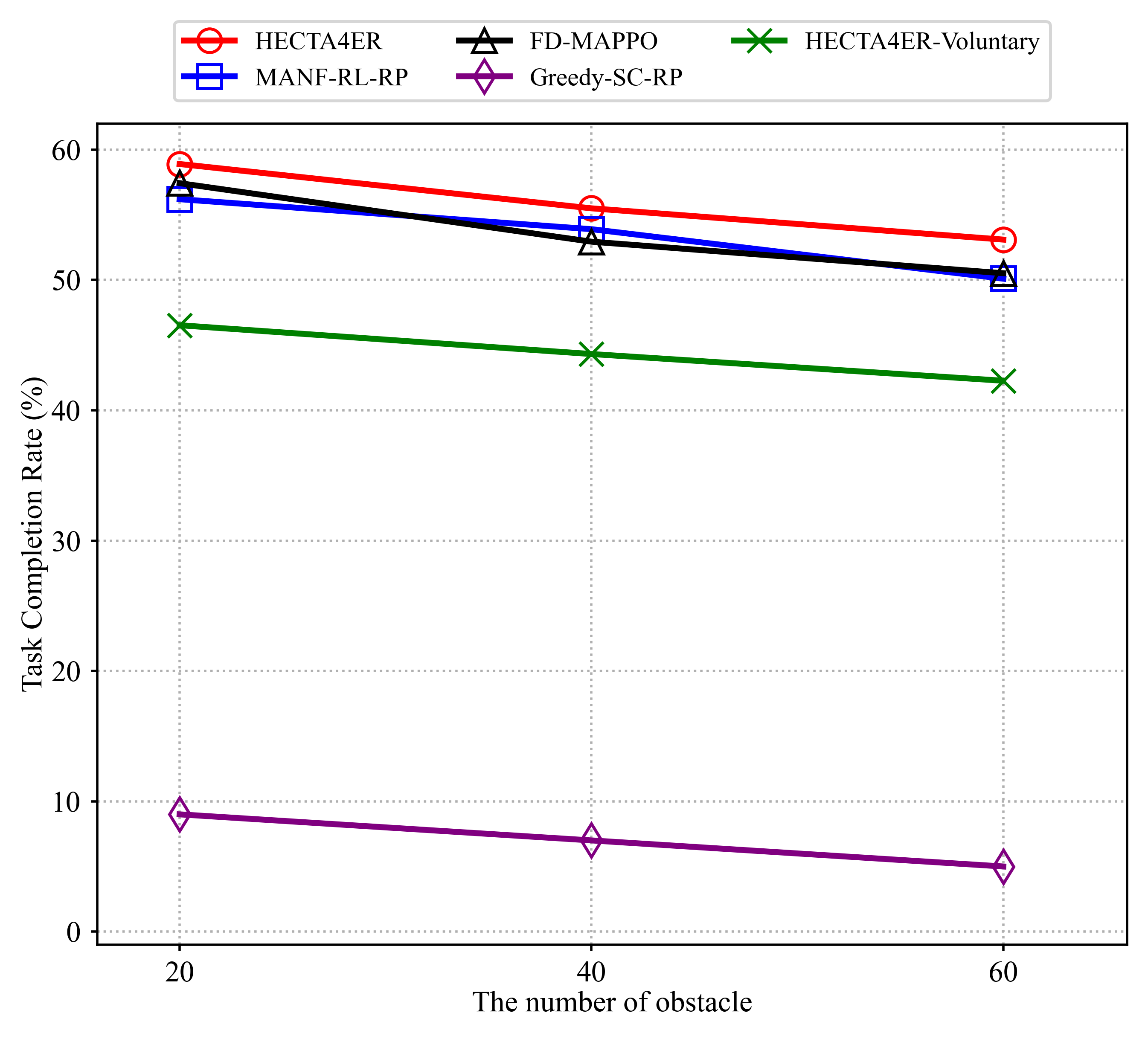}%
\label{Scenario 10}}
\caption{Experimental results of different scenario parameters.}
\label{参数实验}
\end{figure}

\subsubsection{Impact of entity number} We tested how the total number of sensing entities affects performance, keeping the ratio between entity types constant (\(TimeLimit = 9\)). As shown in Fig. \ref{Scenario 5}, increasing the number of entities boosts the TCR for all algorithms. HECTA4ER consistently achieves the highest TCR across all tested quantities.

\subsubsection{Impact of environment size} We tested the impact of varying the environment size with a time limit of 9. As shown in Fig. \ref{Scenario 6}, increasing the sensing area reduces the TCR for all algorithms except Greedy-SC-RP, indicating that larger areas make task completion more difficult. Greedy-SC-RP employs a simple nearest-task greedy strategy. This approach scales well, efficiently finding nearby tasks and maintaining stable performance even as the environment size increases. Conversely, RL algorithms may struggle with the increased exploration required in larger areas, often resulting in a decreasing TCR.

\subsubsection{Impact of task number} We tested the impact of varying the task number, keeping the time limit at 9. As shown in Fig. \ref{Scenario 7}, it reveals opposite trends: TCR decreases for reinforcement learning algorithms but increases for Greedy-SC-RP as the number of tasks increases. This implies a task completion bottleneck within the fixed time, where overloading reduces the RL algorithms' overall effectiveness (TCR). Unlike RL algorithms that aim for long-term optimality, Greedy-SC-RP makes short-sighted decisions by always selecting the nearest task. As the number of tasks increases, the probability of finding a nearby task also increases, enabling the algorithm to complete more tasks within the time limit.

\subsubsection{Impact of task type number} We tested the impact of task type distribution using two settings: a uniform distribution and a distribution proportional to the available sensing entity types (\(TimeLimit = 9\)). Results (Fig. \ref{Scenario 8}) show that the distribution matching entity categories achieves a higher TCR, highlighting the benefit of aligning tasks with specialized sensing entities.

\subsubsection{Impact of task execution time} We evaluated performance using three different distributions of task execution times (\(TimeLimit = 9\)). The results (Fig. \ref{Scenario 9}) indicate that increasing the proportion of short-duration tasks improves the overall TCR, whereas a higher proportion of long-duration tasks reduces it.

\subsubsection{Impact of obstacle number} We varied the number of obstacles in the environment, running experiments with a time limit of 9. As shown in Fig. \ref{Scenario 10}, increasing the obstacle count leads to a decrease in TCR for all tested algorithms.

\subsection{Case study}

\begin{figure}[!t]
\centering
\subfloat[Satellite map of Zhongfu Community]{\includegraphics[width=1.5in]{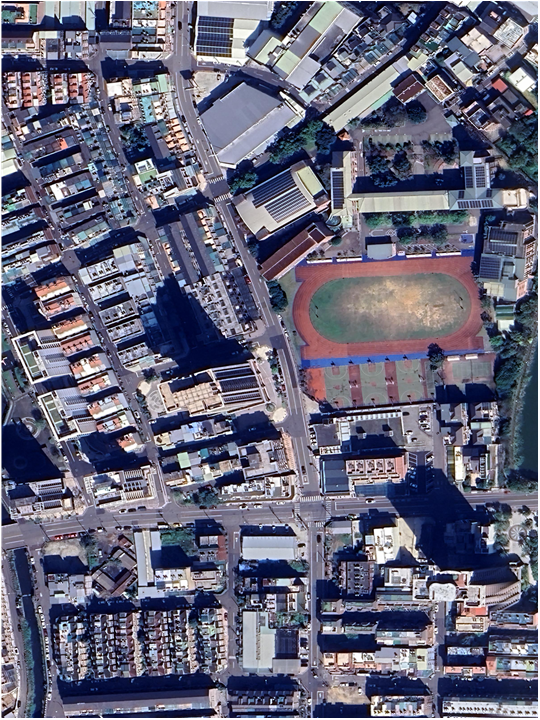}%
\label{中福社区地图}}
\hfil
\subfloat[Visualization of the task scenario]{\includegraphics[width=1.7in]{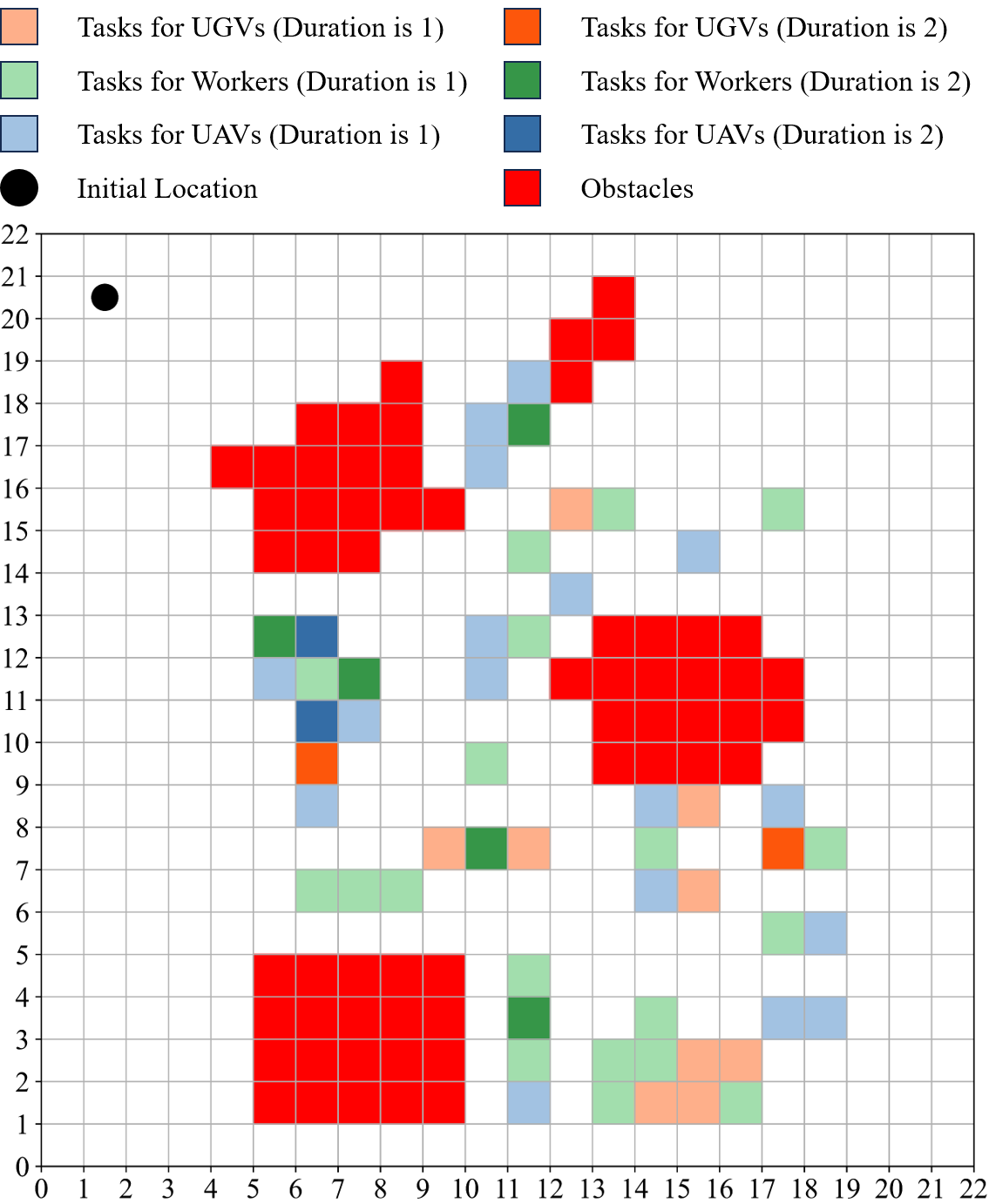}%
\label{中福社区离散展示}}
\caption{Real-world scene of the case study.}
\label{中福社区}
\end{figure}

We utilized the Zhongfu Community in Taoyuan City ($25^{\circ}11^{\prime}0^{\prime\prime}$ N, $121^{\circ}24^{\prime}0^{\prime\prime}$ E), a real-world setting spanning approximately 12 \(\text{km}^2\). Environmental data, including roads and buildings sourced from Baidu Maps, is depicted in the satellite map (Fig. \ref{中福社区地图}). The community's diverse urban infrastructure (schools, residences, sports venues, hospitals) makes it an appropriate testbed for task allocation studies.

The preprocessing is mainly carried out in the following aspects. First, we discretized the Zhongfu Community into a 20x20 grid (400 cells, 170m $\times$ 170m each). Second, obstacles were defined as tall buildings (hindering UAVs) and certain open areas (like playgrounds, not needing sensing). Third, we set 54 tasks: 24 for humans, 11 for UGVs, and 19 for UAVs. 45 tasks require 1 sensing period for execution, while 9 tasks require 2 periods. The distribution of obstacles and tasks is shown in Fig. \ref{中福社区离散展示}. Finally, we configured 6 sensing entities (2 human workers, 2 UGVs, 2 UAVs). The movement radius of a human worker in one sensing period is 3 cells. The movement radius of a UGV in one sensing period is 7 cells, and the power detection radius of a UAV is 10 cells. The movement radius of a UAV in one sensing period is 10 cells, and the power consumption ratio per period is 20\%. Movement radii reflect relative speeds based on the previous study \cite{Zhao2024}.

\begin{table}[htbp]
\centering
\caption{Experimental results of the case study over different time limits (TCR, \%).}
\renewcommand{\arraystretch}{1.5}
\begin{tabular}{cccc}
\hline
\hline
Algorithm & Time-6 & Time-9 &  Time-12 \\
\hline
\centering
HECTA4ER                & 64.81 & 77.78 & 87.04 \\
MANF-RL-RP       & 55.56 & 66.67 & 77.78 \\
FD-MAPPO      & 46.29 & 59.80 & 71.37 \\
Greedy-SC-RP  &  4.00 & 9.00 & 11.00 \\
HECTA4ER-Voluntary  &  38.89 & 46.29 & 51.85 \\
\hline
\hline
\label{真实实验结果}
\end{tabular}
\end{table}

We ran experiments with time limits of 6, 9, and 12, keeping neural network and learning parameters consistent with those in simulation scenario experiments. Results (Table \ref{真实实验结果}) show HECTA4ER performs best across all time limits, with its TCR rising from around 65\% to nearly 87\%. MANF-RL-RP also improves significantly (TCR around 55\% to 77\%) but consistently lags behind HECTA4ER. The TCR of the FD-MAPPO algorithm increases from around 46\% to 71\%. HECTA4ER-Voluntary's performance is considerably lower than standard HECTA4ER, while Greedy-SC-RP has the lowest TCR and shows minimal improvement with more time.

\begin{figure*}[htbp]
\centering
\includegraphics[width=6in]{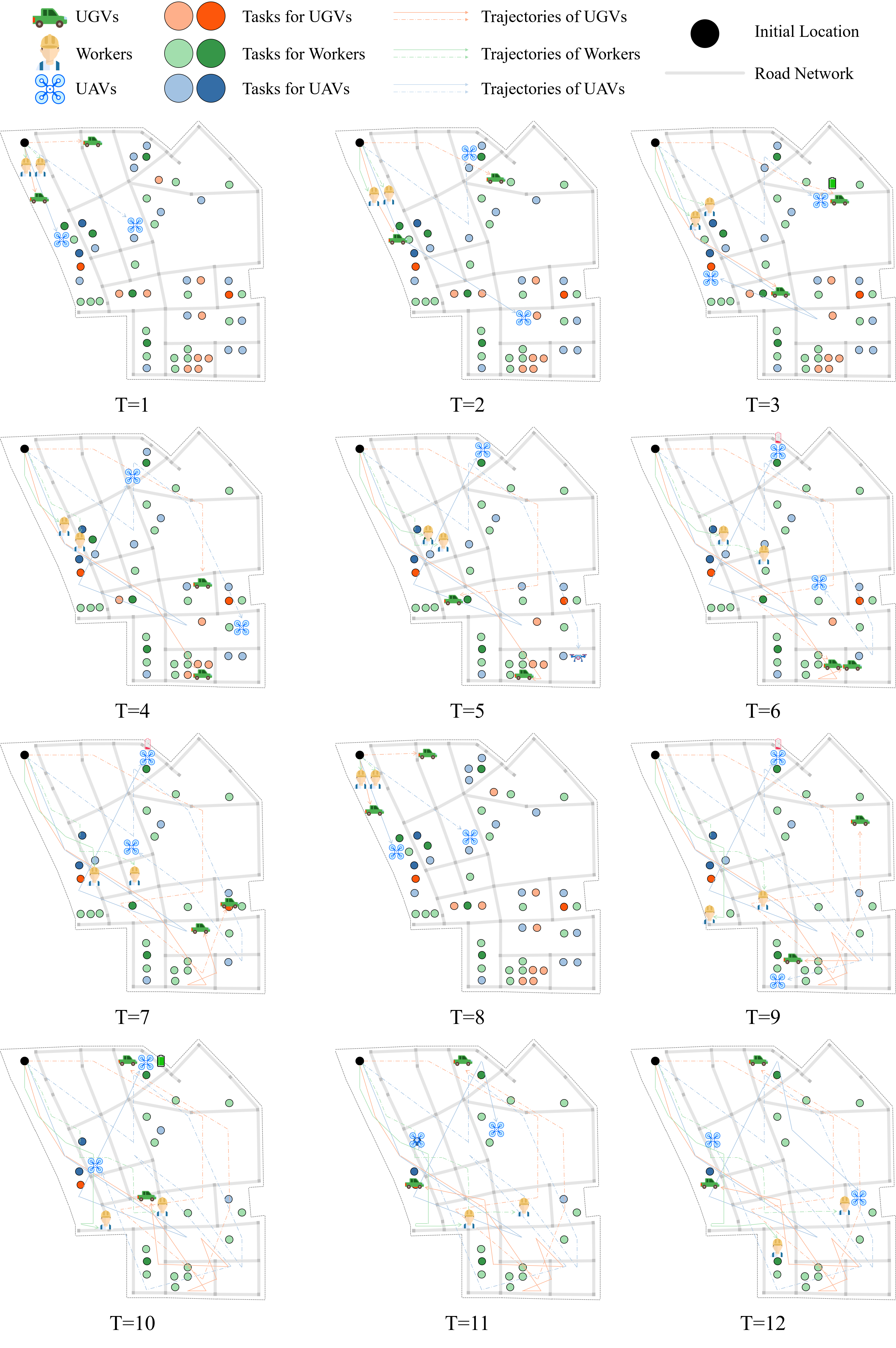}%
\caption{Visualization results of the movement trajectories for human workers, UGVs, and UAVs under the time limit of 12.}
\label{真实实验可视化}
\end{figure*}

We also visualized the movement trajectories of human workers, UGVs, and UAVs under time limit of 12 in Fig. \ref{真实实验可视化}. Analyzing HECTA4ER's trajectories in the Zhongfu Community reveals two key observations. First, we clearly observe the collaboration between UAVs and UGVs in terms of battery replenishment. At times \(T=3\) and \(T=10\), UGVs replenish the batteries of UAVs. Notably, a replenishment fails at \(T=6\) because the depleted UAV was outside the UGVs' detection range. However, at \(T=9\), a UGV proactively detected and replenished a low-battery UAV, enabling it to complete two more tasks – demonstrating the benefit of the ``Hard-Cooperative'' policy. Second, it is observed that sensing entities of the same type tend to distribute themselves among tasks of varying durations (observed for humans at \(T=3\) and UGVs at \(T=6\)), rather than competing for identical tasks or leaving tasks unaddressed.

%% file: 6_discussion.tex
\section{Discussion}\label{6}

\textbf{\textit{Findings.}} By dedicatedly designing a MARL algorithm, this paper addresses the HECTA problem in emergency rescue scenarios. Experimental results convey two main messages: (1) The proposed HECTA4ER algorithm significantly improves task allocation, achieving an average 18.42\% higher TCR compared to baselines while maintaining strong performance in dynamic sensing scenarios. (2) A visualized real-world case study demonstrates effective collaboration among heterogeneous sensing entities and validates HECTA4ER's applicability in complex, real-world environments.

\textbf{\textit{Drawbacks.}} Despite the strengths, this work has several limitations. (1) \textit{Environmental complexity gap:} The use of a 2D grid is an abstraction that cannot fully capture the complexities and continuous nature of real-world 3D environments. (2) \textit{Simplified movement model:} The current movement model only determines target positions for sensing entities, omitting continuous path planning calculations during their transit. (3) \textit{Insufficient decision-making capability:} Existing agents exhibit deficiencies in generalized sensing and reasoning capabilities, resulting in suboptimal performance within dynamic and complex task environments.

\textbf{\textit{Potential avenues.}} Future work will explore the following directions to address these limitations: (1) \textit{Detailed environment modeling:}  Investigate using 3D modeling techniques (e.g., generating point clouds \cite{Liu2023}) and applying semantic annotation/segmentation to create more realistic representations of obstacles and tasks. (2) \textit{Continuous action space and path planning:} Model entity actions within a continuous space and integrate path planning principles (like A*) into the deep reinforcement learning reward function to enable continuous path-finding. (3) \textit{Large model-enabled agents:} To improve the autonomous decision-making capabilities of agents in emergency response, future work could investigate large model-enabled agents and utilize post-training methods to enhance the perception and reasoning capabilities of existing large multimodal models.

%% file: 7_conclusion.tex
\section{Conclusion}\label{7}
In this work, we addressed the critical challenge of optimizing task allocation among heterogeneous sensing entities—humans, UAVs, and UGVs—within the demanding context of emergency rescue operations, characterized by partial observability and operational constraints. We introduced a novel ``Hard-Cooperative'' policy for UAV energy management and formulated the underlying Heterogeneous-Entity Collaborative-Sensing Task Allocation (HECTA) problem as a Dec-POMDP. The proposed MARL-based algorithm, HECTA4ER, built on a CTDE architecture with specialized modules for complex feature extraction and historical information utilization, was designed specifically to tackle this complex, partially observable environment. Our extensive simulations demonstrated HECTA4ER's superiority, achieving a significant average increase of 18.42\% in task completion rate over established baselines. Crucially, a real-world case study confirmed the algorithm's practical effectiveness and robustness in dynamic scenarios. These findings collectively underscore the potential of HECTA4ER as a robust and efficient solution for coordinating heterogeneous multi-agent systems in time-critical emergency response applications.


%% file: Appendix.tex
\appendix

\section{Appendix}

The theoretical proofs involved in this paper are detailed in this appendix.

\subsection{Full cooperativity}
\label{appendix-a}
\textbf{Theorem 1.} \textit{Human workers, UAVs, and UGVs operate in a fully cooperative manner.} 

\textbf{Proof.} All entity types share the common overarching objective of maximizing the number of completed sensing tasks within the time limit. Furthermore, the specific ``Hard-Cooperative'' policy establishes an explicit cooperative interaction between UGVs and UAVs, where UGVs deviate from their sensing tasks to support the UAV battery swapping. Given the shared global objective and defined cooperative interactions, the heterogeneous entities operate under a fully cooperative paradigm. Q.E.D.

\subsection{NP-hard problem}
\label{appendix-b}
\textbf{Theorem 2.} \textit{The HECTA problem is NP-hard.}    

\textbf{Proof.}  We prove the NP-hardness by showing that a known NP-hard problem, the Orienteering Problem (OP), can be reduced to a special case of our problem, which we denote as \textbf{Problem 1}. The reduction strategy involves simplifying \textbf{Problem 1}. Assume that only UAVs can execute sensing tasks and the task execution is not affected by human workers or UGVs, \textbf{Problem 1} can then be expressed as \textbf{Problem 2}:
\allowdisplaybreaks
\begin{align}
\begin{array}{lll}
&{\textbf{confirm}}  & \textbf{DTRA}
\\
&{\textbf{max}} &\frac{N_0-N(TimeLimit)}{N_0}
\\
&\text{s.t.}     &(\ref{公式6}),(\ref{公式7}),(\ref{公式8})
\\
&     &dLoc^t_{k_2}.po=0
\\
&     &dLoc^{t+1}_{k_2}\in dRge^t_{k_2}
\\
&     &N_t=N_{t-1}-1
\\
&     &\textbf{if}\quad\exists k_2, \forall t\in [t,t+\tau dur_i],\ dLoc^t_{k_2}=\tau loc_i
\end{array}
\label{公式17}
\end{align}

\textbf{Problem 2} is a special case of \textbf{Problem 1} and also a variant of the Team Orienteering Problem (TOP), which is known to be NP-hard. If we assume there is only one UAV in the UAV set, \textbf{Problem 2} can be transformed into \textbf{Problem 3}:
\begin{align}
\begin{array}{lll}
&{\textbf{confirm}}  & dTRA_0
\\
&{\textbf{max}} &\frac{N_0-N(TimeLimit)}{N_0}
\\
&\text{s.t.}     &dLoc^t_0.po=0
\\
&     &dLoc^{t+1}_0\in dRge^t_0
\\
&     &po_0+pt_0\leq1
\\
&     &\sum^{N_0}_{i=1}x_{1i}\leq 1   
\\
&     &N_t=N_{t-1}-1
\\
&     &\textbf{if}\ \ \forall t\in [t,t+\tau dur_i],\ dLoc^t_0=\tau loc_i
\end{array}
\label{公式18}
\end{align}

This \textbf{Problem 3} is precisely an instance of the standard \textbf{OP}, which is a well-established NP-hard problem. Therefore, \textbf{Problem 1} is NP-hard. Q.E.D.

\subsection{Markov property}
\label{appendix-c}
\textbf{Theorem 3.} \textit{The belief state satisfies the Markov property.
}

\textbf{Proof.} Referring to Eq. \ref{公式信念空间}, the belief state is the posterior probability that the system is in state $s^t$, as shown in Eq. \ref{公式后验概率}, where $I_k^t = \{o^1_k,a^1_k,...,o^{t-1}_k,a^{t-1}
_k,o^{t}_k,b_0\}$ is $k$-th entity's history information at time $t$.

\begin{align}
    b_k(s^t) = P(s^t|I_k^t) \label{公式后验概率}
\end{align}

According to Bayes' rule, we can represent $b_k(s^{t})$ as Eq. \ref{公式信念状态贝叶斯公式}. $P(o^t_k|s^{t},I_k^{t-1},a^{t-1}_k)$ is the probability of observing $o^t_k$ after taking action $a^{t-1}_k$ in state $s^{t}$, which is the observation function $\Omega(s^{t},a^{t-1}_k,o^t_k)$. $P(s^{t}|I_k^{t-1},a^{t-1}_k)$ is the probability of transitioning to state $s^{t+1}$ after taking action $a^{t-1}_k$, which can be calculated through the state transition probability $T_r(s^{t}|s^{t-1},a^{t-1}_k)$ and the previous belief state $b_{k}(s^{t-1})$. $P(o^t_k|I_t^{t-1},a^{t-1}_k)$ is the normalized factor of observing $o^t_k$, ensuring the sum of probabilities equals 1.

\begin{align}
    \begin{split}
        b_k(s^{t})&=P(s^{t}|I_k^t)\\
        &=\frac{P(o^t_k|s^{t},I_k^{t-1},a^{t-1}_k)\cdot P(s^{t}|I_k^{t-1},a^{t-1}_k)}{P(o^t_k|I_t^{t-1},a^{t-1}_k)}
    \end{split}
    \label{公式信念状态贝叶斯公式}
\end{align}

State transition probability $P(s^{t}|I_k^{t-1},a^{t-1}_k)$ can be represented as Eq. \ref{公式状态转移概率}, where $P(s^{t}|s^{t-1},a^{t-1}_k)$ is the state transition probability $T_r(s^{t}|s^{t-1},a^{t-1}_k)$ and $P(s^{t-1}|I_{t-1})$ is the previous moment belief state $b_{k}(s^{t-1})$.

\begin{align}
    \begin{split}
        P(s^{t}|I_k^{t-1},a^{t-1}_k) &= \sum_{s^{t-1}\in S}P(s^{t}|s^{t-1},a^{t-1}_k)\cdot P(s^{t-1}|I_k^{t-1})\\
        &=\sum_{s^{t-1}\in S} T_r(s^{t}|s^{t-1},a^{t-1}_k)\cdot b_{k}(s^{t-1})
    \end{split}
            \label{公式状态转移概率}
\end{align}

The normalized factor $P(o^t_k|I_k^{t-1},a^{t-1}_k)$ can be represented as Eq. \ref{公式归一化因子}. Substituting the above results into the expression of Bayes' rule, we can obtain Eq. \ref{公式信念状态final}, where $s^{''}$ is a temporary variable used to iterate through all possible next states.

\begin{figure*}[!b]
\hrulefill
    \centering
    \begin{align}
        P(o^t_k|I_k^{t-1},a^{t-1}_k) = \sum_{s^{t}\in S}P(o^t_k|s^{t},a^{t-1}_k) \cdot P(s^{t}|I_k^{t-1},a^{t-1}_k) 
        =\sum_{s^{t}\in S}\Omega(s^{t},a_{k}^{t-1},o^t_k)\cdot \sum_{s^{t-1}\in S} T_r(s^{t}|s^{t-1},a^{t-1}_k) \cdot b_{t-1}(s^{t-1})
    \label{公式归一化因子}
\end{align}

\end{figure*}

\begin{figure*}[!b]
    \centering
    \begin{align}
        b_k(s^t) = \frac{\Omega(s^t, a_k^{t-1}, o_k^t) \cdot \sum_{s^{t-1} \in S} T_r(s^{t}|s^{t-1},a^{t-1}_k) \cdot b_{k}(s^{t-1})}{\sum_{s'' \in S} \Omega(s'', a_k^{t-1}, o_k^t) \cdot \sum_{s^{t-1} \in S} T(s''|s^{t-1}, a_k^{t-1}) \cdot b_{k}(s^{t-1})} 
    \label{公式信念状态final}
\end{align}
\end{figure*}

Observing Eq. \ref{公式信念状态final}, which involves only $b_{k}(s^{t-1})$, $a^{t-1}_k$ and $o^{t}_k$, and does not directly depend on all the belief states before sensing period $t-1$. Moreover, the belief state $b_k(s^{t-1})$ is a sufficient statistic of the historical information $I_{t-1}$, hence we have Eq. \ref{公式马尔科夫性质}, meaning the belief state satisfies the Markov property, where $b_{k,t-1}$ represents the observation probability distribution over all states at sensing period $t-1$.  Q.E.D.

\begin{align}
    P(b_k(s^t)|b_k(s^{t-1}),b_k(s^{t-2}),...,b_k(s_0))=P(b_k(s^t)|b_k(s^{t-1}))\label{公式马尔科夫性质}
\end{align}

\subsection{Convergence analysis}
\label{appendix-d}

\textbf{Theorem 4.} \textit{Algorithm \ref{算法1} converges in finite steps.}

\textbf{Proof.}

    \textbf{Base Case:}  
    At initialization, the individual Q-values \(Q_k(o^0_{c,k},h^0_k, a^0_k)\) are randomly initialized. Assume that \(Q_k(o^0_{c,k},h^0_k, a^0_k)\) satisfies Eq. \ref{公式20}, where \(Q^*_k(o^0_{c,k}, h^0_k,a^0_k)\) is the true optimal Q-value and \(\epsilon_0\) is a positive constant. The difference between the initial Q-value and the optimal Q-value is bounded:  
    \begin{align}
    |Q_k(o^0_{c,k}, h^0_k,a^0_k) - Q^*_k(o^0_{c,k}, h^0_k,a^0_{k})| \leq \epsilon_0  \label{公式20}       
    \end{align}

    \textbf{Inductive Hypothesis:}  
    Assume that at time step \(t\), the Q-values of all entities satisfy Eq. \ref{公式21}, where \(\epsilon_t\) is a positive sequence that decreases over time:  
    \begin{align}
    |Q_k(o^t_{c,k},h^t_k, a^t_k) - Q^*_k(o^t_{c,k},h^t_k, a^t_k)| < \epsilon_t \label{公式21}
    \end{align}

    \textbf{Inductive Step:}  
    We aim to show that at time step \(t+1\), the error is bounded by \(\epsilon_{t+1}\), where \(\epsilon_{t+1} \leq \epsilon_t\) and \(\epsilon_t \to 0\) as \(t \to \infty\)).
    \begin{align}
    |Q_k(o^{t+1}_{c,k},h^{t+1}_k, a^{t+1}_{k}) - Q^*_k(o^{t+1}_{c,k}, h^{t+1}_k,a^{t+1}_k)| < \epsilon_{t+1}     \label{公式22}    
    \end{align}

    The Q-learning update rule is given by Eq. \ref{公式23}. From  Eq. \ref{公式23}, Eq. \ref{公式24} can be derived, where \(\delta_t\) is the Bellman error, and \(\alpha_t\) is the learning rate satisfying the Robbins-Monro conditions: \(\sum \alpha_t = \infty\) and \(\sum \alpha_t^2 < \infty\). 

    \begin{figure*}[!b]
    \begin{align}
    Q_k^{new}(o^{t}_{c,k}, h^{t}_k, a^{t}_k) \leftarrow Q_k(o^t_{c,k},h^{t}_k, a^t_k) + \alpha_t [ r^t + \gamma \max_{a^{\prime,t}_k} Q_k(o^{t+1}_{c,k}, h^{t+1}_k,a^{\prime,t}_k)  
         - Q_k(o^t_{c,k},h^{t}_k, a^t_k) ]         \label{公式23}
    \end{align}
    \begin{align}
        Q_k^{new}(o^{t}_{c,k}, h^{t}_k, a^{t}_k) &= (1-\alpha_t)Q_k(o^t_{c,k}, h^{t}_k, a^t_k) + \alpha_t [r_t + \gamma \max_{a^{\prime}_k} Q_k(o^{t+1}_{c,k}, h^{t+1}_k, a^{\prime}_k)] \nonumber \\
        &= Q_k(o^t_{c,k}, h^{t}_k, a^t_k) + \alpha_t \delta_t  \label{公式24}
    \end{align}
    
    \end{figure*}


    Eq. \ref{公式25} is the Bellman equation, where \(R\) is the expected immediate reward. The Bellman error \(\delta_t\) is an estimate of the difference between the current Q-value and the target value derived from the Bellman equation. Under suitable conditions, the expected Bellman error \(\mathbb{E}[\delta_t]\) approaches zero as \(Q_k \to Q_k^*\), and its variance is bounded.


    \begin{align}
        \begin{split}
            Q_k^*(o_{c,k}^t, h^{t}_k, a_k^t) = \mathbb{E}&_{o_{c,k}^{t+1}, h_k^{t+1} \sim P} [ R(o_{c,k}^t, a_k^t) \\ 
            &+ \gamma \max_{a_k^{\prime}} Q_k^*(o_{c,k}^{t+1}, h_k^{t+1}, a_k^{\prime}) ] \label{公式25} 
        \end{split}
    \end{align}


    After incorporating the mixing module, the global value function is expressed as in Eq. \ref{公式27}. This module adjusts individual Q-values using the joint action value network \(Q_{tot}\) and the environmental state value function \(V(s^{t}_c, H^t)\), ensuring consistency and coordination in the overall strategy. Even if some entities take non-optimal actions, the overall performance remains unaffected as long as other entities compensate for the loss.

    \begin{align}
        \begin{split}
         \hat{Q}_{tot}(s^t_c, H^t, A^t) &= Q_{tot}(s^t_c, H^t, A^t) \\ 
         &+ V(s^t_c, H^t) - \sum_k Q_k(s^t_c, h^{t}_k,a_k^t) \label{公式27}
        \end{split}
    \end{align}

     Since Q-values are updated following the Bellman equation and the mixing module provides a global value assessment, Eq. \ref{公式28} holds. According to the Robbins-Monro conditions, the step size parameter \(\alpha\) ensures that cumulative errors gradually diminish, eventually approaching zero. Thus, Eq. \ref{公式29} is satisfied, where \(\epsilon_{t+1} < \epsilon_t\).

    \begin{align}
        \begin{split}
        &|Q_k(o^{t+1}_{c,k},h^{t+1}_k, a^{t+1}_k) - Q^*_k(o^{t+1}_{c,k},h^{t+1}_k,  a^{t+1}_k)| \leq \; \\& (1 - \alpha_t) |Q_k(o^t_{c,k},h^{t}_k,  a^t_k) - Q^*_k(o^{t+1}_{c,k}, h^{t+1}_k, a^{t+1}_k)|  + \alpha_t |\delta_t| \\
        &=(1-\alpha_t)\epsilon_t+\alpha_t|\delta_t|   \label{公式28}
        \end{split}
    \end{align}

    \begin{align}
        \begin{split}
        |Q_k(o^{t+1}_{c,k},h^{t+1}_k,  a^{t+1}_{k}) - Q^*_k(o^{t+1}_{c,k},h^{t+1}_k,  a^{t+1}_k)| < \epsilon_{t+1} \label{公式29}
        \end{split}
    \end{align}

     Given that the variance of the Bellman error is bounded, an appropriate step size \(\alpha_t\) can be chosen to satisfy Eq. \ref{公式30}, keeping the Bellman error within an acceptable range. \(\sigma\) is a normalizing constant.

    \begin{align}
        \alpha_t \leq \frac{\epsilon_t}{\sigma} \label{公式30}
    \end{align}

     A recursive relationship can be established as in Eq. \ref{公式31}. To ensure \(\epsilon_{t+1} < \epsilon_t\), Eq. \ref{公式32} must hold. Since \(\sigma > 0\) and \(\epsilon_t > 0\), an appropriate \(\alpha_t\) can be selected to satisfy this condition.  
     \begin{align}
             \epsilon_{t+1} \leq (1 - \alpha_t) \epsilon_t + \alpha_t \sigma \label{公式31}
     \end{align}
     \begin{align}
         (1 - \alpha_t) \epsilon_t + \alpha_t \sigma < \epsilon_t \implies \alpha_t (\sigma - \epsilon_t) < 0 \label{公式32}
     \end{align}

     According to the Robbins-Monro conditions, the step size is chosen as \(\alpha_t = \frac{C}{t^\beta}\), where \(C > 0\) and \(0 < \beta < 1\). This ensures the step size diminishes over time while the cumulative step size diverges, guaranteeing algorithm convergence within a finite number of steps. Q.E.D.